\documentclass{ecai} 

\usepackage{latexsym}
\usepackage{amssymb}
\usepackage{amsmath}
\usepackage{amsthm}
\usepackage{booktabs}
\usepackage{enumitem}
\usepackage{graphicx}
\usepackage{color}
\usepackage{algorithm}
\usepackage{algorithmic}
\usepackage{subcaption}
\usepackage{caption}

\newtheorem{theorem}{Theorem}
\newtheorem{lemma}[theorem]{Lemma}

\newtheorem{proposition}[theorem]{Proposition}

\newtheorem{assumption}[theorem]{Assumption}
\newtheorem{definition}[theorem]{Definition}
\newtheorem{remark}[theorem]{Remark}

\newcommand{\BibTeX}{B\kern-.05em{\sc i\kern-.025em b}\kern-.08em\TeX}

\begin{document}


\begin{frontmatter}


\paperid{6810} 


\title{Degree of Staleness-Aware Data Updating \\
in Federated Learning}


\author{\fnms{Tao}~\snm{Liu}}
\author{\fnms{Xuehe}~\snm{Wang}\thanks{Corresponding Author. Email: wangxuehe@mail.sysu.edu.cn.}}

\address{School of Artificial Intelligence, Sun Yat-sen University}


\begin{abstract}
Handling data staleness remains a significant challenge in federated learning with highly time-sensitive tasks, where data is generated continuously and data staleness largely affects model performance. Although recent works attempt to optimize data staleness by determining local data update frequency or client selection strategy, none of them explore taking both data staleness and data volume into consideration. In this paper, we propose DUFL(\underline{\textbf{D}}ata \underline{\textbf{U}}pdating in \underline{\textbf{F}}ederated \underline{\textbf{L}}earning), an incentive mechanism featuring an innovative local data update scheme manipulated by three knobs: the server's payment, outdated data conservation rate, and clients' fresh data collection volume, to coordinate staleness and volume of local data for best utilities. To this end, we introduce a novel metric called DoS(the \underline{\textbf{D}}egree \underline{\textbf{o}}f \underline{\textbf{S}}taleness) to quantify data staleness and conduct a theoretic analysis illustrating the quantitative relationship between DoS and model performance. We model DUFL as a two-stage Stackelberg game with dynamic constraint, deriving the optimal local data update strategy for each client in closed-form and the approximately optimal strategy for the server. Experimental results on real-world datasets demonstrate the significant performance of our approach.
\end{abstract}

\end{frontmatter}

\section{Introduction}
\label{Introduction}
With the development of Internet of Things (IoT), billions of end devices are generating plenty of data each day. 
The traditional approach of uploading data to a remote cloud for model training may lead to various problems including privacy concern and communication burden.
As a privacy-preserving machine scheme, federated learning provides a solution by supporting clients jointly training the global model while keeping data local. 
It has been widely applied in numerous scenarios such as smart city \cite{al2023federated}, autonomous vehicles \cite{zeng2022federated} and personalized recommendation \cite{pfitzner2021federated}.
 
Despite various advantages, federated learning still faces two bottlenecks:
1) \textit{Staleness of Data:} In most existing works, the clients are assumed to hold a static dataset and use the same dataset to train the local model over the time horizon.
But in reality, there are many highly time-sensitive tasks with streaming data, where data is generated continuously and becomes out-of-date over time \cite{xiao2023aoi}.
Under this circumstance, outdated data used for training may deteriorate model parameters and reduce service quality.
2) \textit{Training Cost:} When participating in model training, clients consume various resources such as computing capacity and communication bandwidth.
Besides, collecting fresh data frequently to meet the requirement of highly time-sensitive tasks also causes extra cost. Considering the huge expenditure of resources, clients may be reluctant to provide fresh data or even refuse to participate in federated learning tasks without economic compensation.

Efforts have been made in incentive mechanism design ranging from game theory \cite{lim2020hierarchical,zhang2022enabling,huang2024collaboration} to auction theory \cite{jiao2020toward} and contract theory \cite{kang2019incentive,ding2020optimal}. However, most of these studies don't pay attention to the negative impact of outdated data on model performance.
A few works have studied Age-of-Information awared incentive mechanisms in mobile crowdsensing \cite{xiao2023aoi,wang2021taming} and federated learning field \cite{wang2019dynamic,wu2023towards}. However, these approaches primarily focus on controlling the data update frequency, without accounting for the combined effect of data staleness and data volume on the overall model training performance.

Motivated by the above discussions, this paper introduces an innovative dynamic local data update scheme with incentive mechanism design, called DUFL, by considering the influences of data staleness and data volume on the model performance and strategy determination.
There are three key challenges in DUFL.
First, \textit{complexity of optimization problem}. The server needs to jointly decide the payment to encourage clients to collect fresh data and the conservation rate to force clients to abandon outdated data, while clients face a long-term optimization problem with dynamic constraints of data update. Solving the above two interdependent optimization problems is challenging due to their mutual influence on each other.
Second, \textit{absence of a quantitative model performance characterization in DUFL}. Although model performance is influenced by data staleness and data volume, there's lack of quantitative relationship linking these factors to model performance, which impedes the strategy determination for the server.
Third, \textit{incomplete information for strategy determination}. A client's optimal data update strategy is interdependent with others' strategies, as the allocated payment is based on relative contribution. But in real-world scenarios, the lack of communication among clients obscures others' private decisions, making individual strategy optimization challenging.

To overcome the above challenges, we introduce a new metric to measure the data staleness, and leverage it to conduct a theoretic analysis for DUFL. The main contributions in this paper are summarized as follows:
\begin{itemize}
  \item We propose a new framework called DUFL(\underline{\textbf{D}}ata \underline{\textbf{U}}pdating in \underline{\textbf{F}}ederated \underline{\textbf{L}}earning) for highly time-sensitive federated learning tasks. The central server decides the payment and the conservation rate to balance monetary payment and model accuracy loss. Each client dynamically controls fresh data collection volume based on the server's strategy to balance its costs and allocated payment.
  \item We introduce a new metric called DoS(the \underline{\textbf{D}}egree \underline{\textbf{o}}f \underline{\textbf{S}}taleness) to measure the data staleness. Based on it, we conduct theoretic analysis and obtain a quantitative relationship linking model performance with data volume and data staleness.
  \item We model DUFL as a two-stage Stackelberg game with dynamic constraints. By means of backward reduction, we derive clients' optimal data update strategy via Hamilton equation and the server's approximately optimal strategy by adopting Bayesian Optimization.
  \item We provide justifications for the effectiveness of DUFL in the respect of both utility and accuracy in comparison with other benchmarks. 
\end{itemize}

\section{Related Work}
\label{Related Work}
\subsection{Incentive Mechanism}
Incentive mechanism designed for federated learning has been widely investigated in previous works.
They can be assorted into three categories:
(1) \textit{game theory:} 
\cite{lim2020hierarchical}  proposes a hierarchical incentive mechanism based on coalitional game theory approach, where multiple workers can form various federations.
\cite{zhang2022enabling} builds a incentive mechanism utilizing repeated game theory to enable long-term cooperation among participants in cross-silo federated learning.
(2) \textit{auction theory:} 
\cite{jiao2020toward} designs two auction mechanisms for the federated learning platform to maximize the social welfare of the federated learning services market.
(3) \textit{contract theory:} 
\cite{kang2019incentive} proposes an effective incentive mechanism combining reputation with contract theory to motivate high-reputation mobile devices with high-quality data to participate in model learning.
\cite{ding2020optimal} presents an analytical study on the server's optimal incentive mechanism design by contract theory, in the presence of users' multi-dimensional private information.
However, the above studies don't take into consideration the negative impact of outdated data on model performance and they can't be applied in highly time-sensitive tasks.

\subsection{Data Staleness Optimization}
Many efforts have been devoted to data staleness optimization.
\cite{tripathi2021age} consider the problem of minimizing AoI(age of information) in general wireless networks.
\cite{fang2021computing} devises a joint preprocessing and transmission policy to minimize the average AoI and the energy consumption at the IoT device.
Nevertheless, only a few works among them study the AoI optimization with economic consideration. \cite{xiao2023aoi} investigates the incentive mechanism design in MCS systems that take the freshness of collected data and social benefits into concerns.
\cite{wang2021taming} considers a general multi-period status acquisition system, aiming to maximize the aggregate social welfare and ensure the platform freshness.
However, they can't be applied in the context of federated learning considering the resource consumption against local model training.
In the federated learning field, \cite{wang2019dynamic} proposes dynamic pricing for the server to offer age-dependent monetary returns and encourages clients to sample information at different rates over time.
\cite{wu2023towards} aims to minimize the loss of global model for FL with a limited budget by determining a client selection strategy under time-sensitive scenarios.
But none of them explore to manipulate local data update by decaying outdated data and collecting fresh data simultaneously considering data staleness and data volume together. 

\section{Problem Formulation}
\subsection{Federated Learning with Data Update}
\begin{figure}[ht]
  \centering
  \includegraphics[width=\columnwidth]{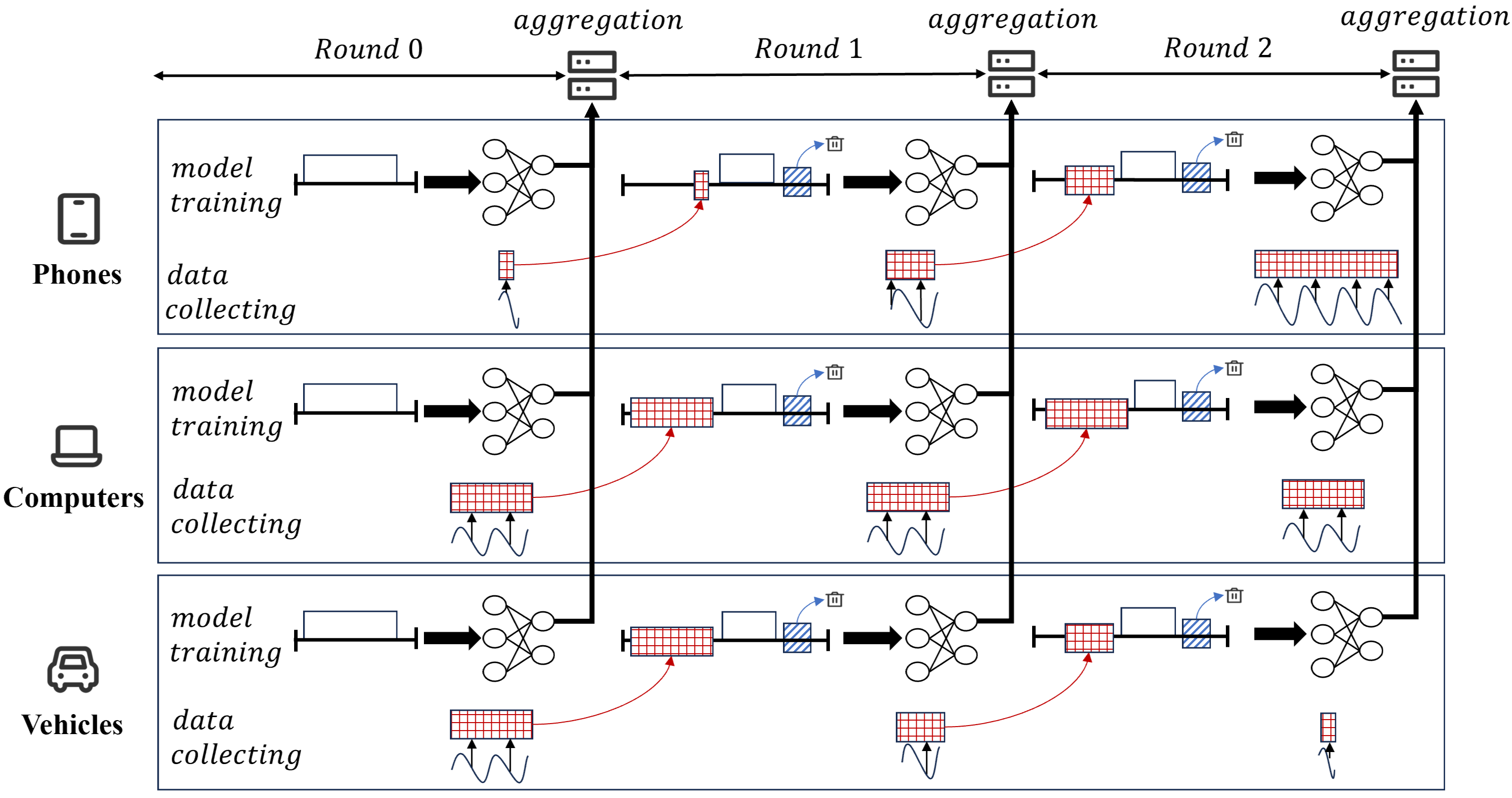}
  \caption{The framework of buffered data update scheme. Tilde denotes local data stream. Blank blocks symbolize buffered data, while red blocks and blue blocks symbolize new data to be collected and stale data to be decayed respectively.}
  \label{fig:framework}
  \vspace{1.5em}
\end{figure}
We assume that there is a central server and $N$ clients in the federated learning system and they are arranged to conduct $T$ rounds of global training. Different from the static local dataset in previous works, each client $k$ has a local data stream $\mathcal{D}_k$ which generates fresh data continuously over the time horizon.
In addition, due to the storage limit, each client maintains a buffer to store data collected from its local data stream $\mathcal{D}_k$ for model training.
We denote the stored data in the buffer of client $k$ at round $t$ as $\mathcal{D}_k(t)$ with data volume $\vert\mathcal{D}_k(t)\vert = D_k(t)$. 
During each round, the buffer of client $k$ can be updated by decaying part of outdated data and collecting fresh data from its local data stream, then used for local model training.
Therefore, the data update scheme of client $k$ can be formulated as
\begin{equation}
  D_k(t+1) = \theta D_k(t) + \Delta_k(t),
  \label{formulation:update}
\end{equation}
where $\theta \in [0,1]$ is the outdated data conservation rate and decides how much buffered data can be conserved. $\Delta_k(t)$ is the fresh data collection volume by client $k$ at round $t$.

Note that data collection cannot be completed immediately, which means the traditional ``collecting first and then training'' paradigm may lead to severe time latency for federated learning tasks.
To improve task efficiency, in our framework, data collection is designed to proceed in parallel with model training, and fresh data collected during round $t$ will be used to update the buffer and train the model at round $t+1$, as depicted in Figure \ref{fig:framework}.

The loss function of client $k$ based on the global model parameter $w(t)$ using stored data $\mathcal{D}_k(t)$ at round $t$ can be represented as 
\begin{equation}
  F_k(w(t)|\mathcal{D}_k(t)) = \frac{1}{D_k(t)} \sum_{j=1}^{D_k(t)} f(w(t)| x_k^j),
\end{equation}
where $f(w(t)| x_k^j)$ is the loss function of each data point $\{x_k^j, y_k^j\} \in \mathcal{D}_k(t)$.
Then client $k$ updates its local model by
\begin{equation}
  w_k(t+1) = w(t) - \eta \nabla F_k(w(t)| \mathcal{D}_k(t)),
\end{equation}
where $\eta$ is the learning rate and $\nabla F_k(w(t)| \mathcal{D}_k(t))$ is the loss gradient of client $k$ at round $t$. 
Until each client completes its local update and sends $w_k(t+1)$ to the server, it will aggregate them by
\begin{equation}
  w(t+1) = \sum_{k=1}^{N} \frac{D_k(t)}{D(t)} w_k(t+1),
\end{equation}
where $\mathcal{D}(t) = \cup_{k=1}^N \mathcal{D}_k(t)$ and $D(t) = \sum_{k=1}^{N} D_k(t)$.
Subsequently, the server launches the new global model $w(t+1)$ to each client for the next round's training.
Afterwards, the global loss function over the time horizon is defined as
\begin{equation}
  F(w) \triangleq \sum_{t=0}^{T-1} \sum_{k=1}^{N} \frac{D_k(t)}{D(t)} F_k(w(t)| \mathcal{D}_k(t)),
\end{equation}
which is similar to \cite{wu2023towards}.
The goal of federated learning is to find the optimal parameters $w^*$ to minimize $F(w)$ as
\begin{equation}
  w^* = \arg \min_{w} F(w).
\end{equation}

\subsection{Convergence Analysis for Federated Learning with Data Stream}
Convergence analysis for model performance is provided in this section.
In practice, it's difficult to measure the model performance of DUFL directly due to data dynamics and so on.
Therefore, we provide a convergence upper bound for DUFL, considering the impact of data volume and data staleness on model performance.
Before that, we introduce the concept of Degree of Staleness (DoS) for buffered data as follows.
\begin{definition}
  \label{definition:1}
  The degree of staleness (DoS) $S_k(t)$ for client $k$'s buffered data at round $t$ is defined as 
  \begin{equation}
    \resizebox{\columnwidth}{!}{$
      S_k(t) = 
      \begin{cases}
        \dfrac{\theta D_k(t-1)}{D_k(t)}(S_k(t-1) + 1) + \dfrac{\Delta_k(t-1)}{D_k(t)}, & t > 0; \\
        1, & t = 0,
      \end{cases}
    $}
  \end{equation}
  which is the weighted sum of the conserved data's DoS and the new data's DoS, where $\frac{\theta D_k(t-1)}{D_k(t)}$ and $\frac{\Delta_k(t-1)}{D_k(t)}$ are the ratio of conserved data and new data for model training at round $t$, respectively.
  The conserved data's DoS at round $t$ increases from $S_k(t-1)$ at round $t-1$ to $S_k(t-1) + 1$ at round $t$, while new collected data's DoS is $1$.
\end{definition}

\begin{remark}
  $S_k(t)$ increases with conservation rate $\theta$ and decreases with new collected data volume $\Delta_k(t)$, which reflects the fact that less conserved buffered data and more fresh data contribute to DoS reduction.
\end{remark}
\begin{lemma}
  \label{lemma:generalformula}
  The general formula of $S_k(t)$ can be further derived as
  \begin{equation}
    S_k(t) = \sum_{\tau=0}^{t} \frac{\theta^{t-\tau} D_k(\tau)}{D_k(t)}.    
  \end{equation}
\end{lemma}
The detailed proof is provided in Appendix \ref{proof:generalformula} in the supplementary material.

Moreover, we introduce some assumptions on local loss function $F_k(w|\mathcal D_k(t))$ before the presentation of convergence analysis, which have been widely used in previous works \cite{yu2018parallel, li2019convergence, qu2021federated}.
\begin{assumption}
\label{assumption:4}
For $t \in \{0, \cdots, T - 1\}, k \in \{1, \cdots, N\}$, $F_k(w|\mathcal D_k(t))$ is $\rho$-Lipschitz, i.e., $\forall w_1, w_2, F_k(w_1|\mathcal D_k(t)) - F_k(w_2|\mathcal D_k(t)) \leq \rho \Vert w_1 - w_2 \Vert_2$.
\end{assumption}

\begin{assumption}
\label{assumption:5}
For $t \in \{0, \cdots, T - 1\}, k \in \{1, \cdots, N\}$, $F_k(w|\mathcal D_k(t))$ is $\beta$-Lipschitz smooth, i.e., $\forall w_1, w_2, \Vert\nabla F_k(w_1|\mathcal D_k(t)) - \nabla F_k(w_2|\mathcal D_k(t))\Vert \leq \beta \Vert w_1-w_2 \Vert_2$.
\end{assumption}

\begin{assumption}
\label{assumption:6}
For $t \in \{0, \cdots, T - 1\}, k \in \{1, \cdots, N\}$, $F_k(w|\mathcal D_k(t))$ is $\mu$-strong convex, i.e., $\forall w, F_k(w|\mathcal D_k(t))$ satisfies $F_k(w|\mathcal D_k(t)) - F_k(w^*) \leq \frac{1}{2\mu}\Vert\nabla F_k(w|\mathcal D_k(t))\Vert_2^2$.
\end{assumption}

\begin{assumption}
\label{assumption:7}
Given buffered data $\mathcal D_k(t)$, $\nabla F_k(w(t)|\mathcal D_k(t))$ is unbiased and variance-bounded, i.e.,  
$\mathbb E[\nabla F_k(w(t) | \mathcal D_k(t))]=\nabla F_k(w(t) | \mathcal D_k)$ and  
$\mathbb E\|\nabla F_k(w(t) | \mathcal D_k(t))-\nabla F_k(w(t) | \mathcal D_k)\|^2\le\varsigma_k^2,$ 
where $\varsigma_k^2:= \frac{\psi^2}{D_k(t)} + S_k(t)\sigma^2$, where $\psi$ is a constant, and $\sigma$ is a constant that measures time sensitivity of the server's task.
\end{assumption}

Then, the convergence analysis with DoS is given as follows.
\begin{theorem}
  \label{theorem:upperbound}
  Under Lemma \ref{lemma:generalformula} and Assumptions \ref{assumption:4}-\ref{assumption:7}, with $\eta \leq \frac{1}{2\beta}$, the convergence upper bound after $T$ rounds can be formulated as
    \begin{align}
      & \mathbb E[F(w(T)|\mathcal D(T)) - F(w^*)] \notag \\
      \leq & \underbrace{\vphantom{\sum_{min}^{max}\frac{1}{2}} \kappa_1^{T} \mathbb E[F(w(0)|\mathcal D(0))-F(w^*)]}_{(1)} \notag \\
      & + \sum_{t=0}^{T-1} \kappa_1^{T-1-t} \left[\underbrace{\vphantom{\sum_{min}^{max}\frac{1}{2}} \kappa_2 \frac{N\psi^2}{D(t)}}_{(2)} + \underbrace{\vphantom{\sum_{min}^{max}\frac{1}{2}} \kappa_3 \sum_{k=1}^{N} \frac{D_k(t)}{D(t)} S_k(t) \sigma^2}_{(3)} + \underbrace{\vphantom{\sum_{min}^{max}\frac{1}{2}} \Omega_{t}}_{(4)}\right], 
      \label{formula:upperbound}
    \end{align}
  where $\Omega_t = F(w(t+1)|\mathcal D(t+1)) - F(w(t+1)|\mathcal D(t))$, $\kappa_1 = 1 + 4\mu\beta\eta^2 - 2\mu\eta$, $\kappa_2 = 2\beta\eta^2$, $\kappa_3 = \beta\eta^2$. 
\end{theorem}

The detailed proof is provided in Appendix \ref{proof:upperbound} in the supplementary material.

\begin{remark}
  In Equ. (\ref{formula:upperbound}), terms (2) and (3) measures the influence of data volume and staleness on model performance respectively. Intuitively, the larger and fresher the buffered data is, the better the global model performs. However, discarding more stale data reduces staleness but also decreases the overall data volume. Thus, there exists a trade-off between data volume and staleness.
  In addition, term (3) shows that highly time-sensitive tasks with larger $\sigma$ are affected more heavily by data staleness. Term (4) of $\Omega_t$ captures the expected difference in global loss function by leveraging the buffered data at current round $\mathcal{D}(t+1)$ and at previous round $\mathcal{D}(t)$, reflecting the influence of data fluctuation on model performance.
\end{remark}

\section{Two-Stage Stackelberg Game with DoS}
\label{section:system model}
In this section, we consider the strategies of the server and clients in DUFL.
First, we construct the utility functions and corresponding optimization problems for the server and clients in Sections \ref{section:cost function} and \ref{section:utility function}, respectively.
Then we model them as a two-stage Stackelberg game with dynamic constraint of local data update in Section \ref{section: game}.
\begin{figure}[ht]
  \centering
  \includegraphics[width=\columnwidth]{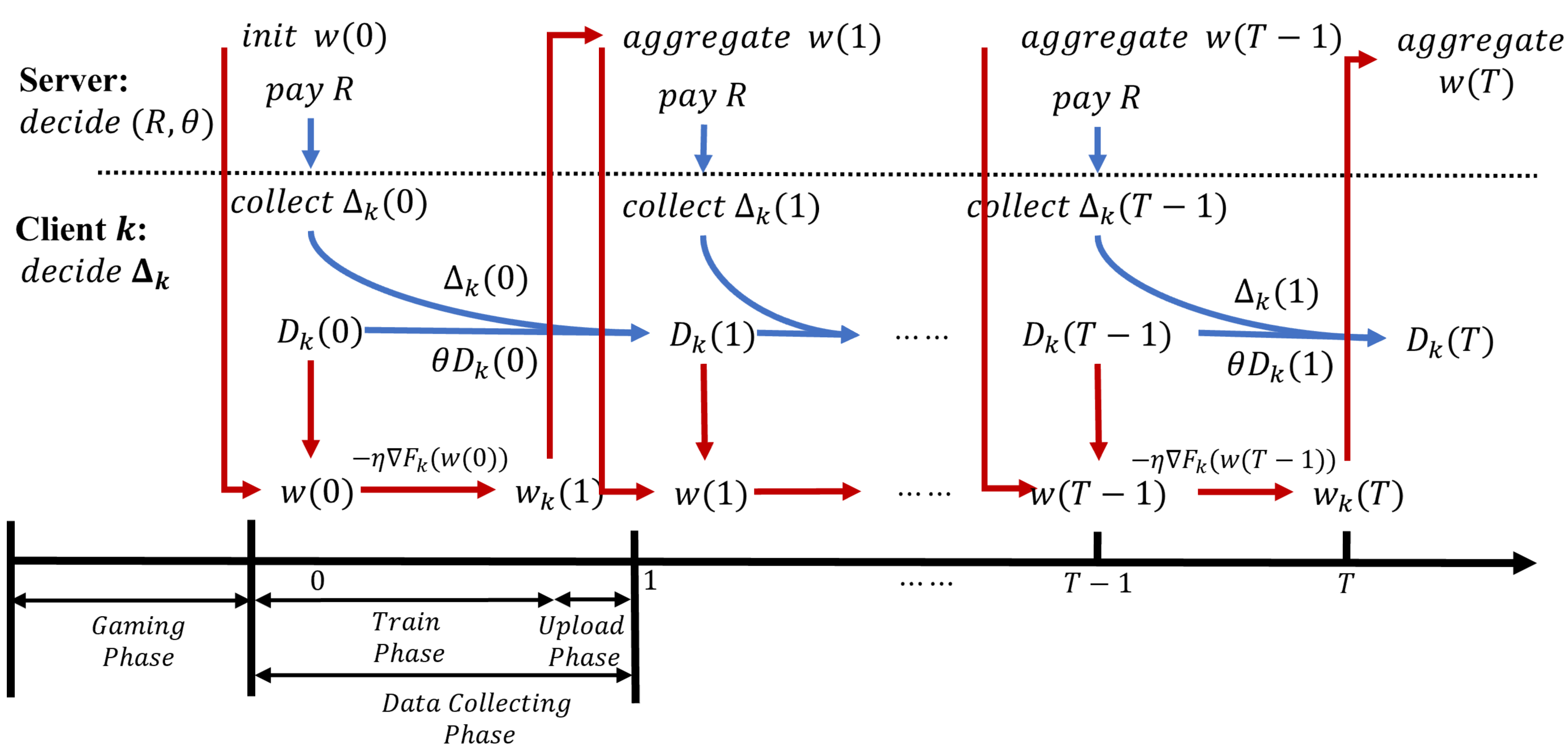}
  \caption{The overview of the decision process in DUFL. Red lines denote model parameter flow while blue lines symbolize information and data flow. The strategies of both parties are agreed upon during gaming phase in advance of training.}
  \label{fig:overview}
\end{figure}

\subsection{Cost Function of Server}
\label{section:cost function}
The cost of the server comprises two parts: the model accuracy loss and the payment to clients. 
In reality, it's difficult to obtain the exact model accuracy loss form directly. Yet we can approximate it with the convergence upper bound provided in Equ. (\ref{formula:upperbound}).
Note that terms $(1)$ and $(4)$ in Equ. (\ref{formula:upperbound}) can't be controlled by the server, thus we extract terms $(2)$ and $(3)$ to evaluate the model accuracy loss.
Denote $R$ as payment to clients per round, and $\theta$ as the outdated data conservation rate. The cost function of the server can be constructed as
\begin{align}
  \label{formulation:cost}
  U(\boldsymbol{\Delta}, R, \theta) & = \sum_{t=0}^{T-1} \Bigg[ \gamma R + (1-\gamma)\kappa_1^{T-1-t} \times \\
  & \bigg(\kappa_2 \frac{N\psi^2}{D(t)} + \kappa_3 \sum_{k=1}^{N} \frac{D_k(t)}{D(t)} S_k(t) \sigma^2 \bigg) \Bigg], \notag
\end{align}
where $\boldsymbol{\Delta} = \{\boldsymbol{\Delta_1}, \cdots, \boldsymbol{\Delta_N}\}$ is clients' data update strategies with client $k$'s strategy denoted as $\boldsymbol{\Delta_k} = \{\Delta_k(0), \cdots, \Delta_k(T-1)\}$, $\gamma \in [0, 1]$ is a factor to balance model performance and cost payment. When $\gamma$ approaches $0$, the server pays more attention to model performance.

Note that the clients' optimal data update strategies $\boldsymbol{\Delta} = \{\boldsymbol{\Delta_1}, \cdots, \boldsymbol{\Delta_N}\}$ are determined based on the payment $R$ and conservation rate $\theta$.
Then, the objective of the server is to determine $R$ and $\theta$ to minimize its cost as
\begin{equation}
  \label{formulation:mincost}
  \min_{R,\theta}  U(\boldsymbol{\Delta}, R, \theta), 
\end{equation}
where a greater reward $R$ leads to more and fresher data for model training but increases monetary payment, while a lower conservation rate $\theta$ forces clients to abandon more stale data but decreases data volume at the same time. This joint optimization problem is non-trivial considering the clients' dynamic data update strategies.

\subsection{Utility Function of Clients}
\label{section:utility function}
A client participating in the federated learning task needs to allocate various resources for data sampling and model training. 
We use $\alpha_k F_1(\Delta_k(t))$ and $\beta_k F_2(D_k(t))$ to depict the expenditure for data collecting and training respectively, where $\alpha_k$ is the unit cost for data collecting and $\beta_k$ is the unit cost for model training.
Both $F_1(\cdot)$ and $F_2(\cdot)$ are convex functions to capture the fact that a client's sampling and training expenditures increase convexly with the data volume.
Here, we adopt the quadratic forms of $F_1(\Delta_k(t)) = \Delta_k^2(t)$ and $F_2(D_k(t)) = D_k^2(t)$, which have been commonly employed in cost formulation \cite{zhan2019free,cheung2017make,nie2020multi}.
Based on this, client $k$'s cost can be expressed as
\begin{equation}
  C_k(t) = \alpha_k \Delta_k^2(t) + \beta_k D_k^2(t).
\end{equation}

Moreover, to motivate clients to participate in model training and encourage them to provide more data, the reward a particular client can receive depends on the size of its buffered data compared with that of other participating clients, which is formulated as
\begin{equation}
\label{formulation:Pkt}
  P_k(t) = \frac{D_k(t)}{\sum_{i=1}^{N}D_i(t)}R.
\end{equation}
Therefore, the utility of a particular client $k$ over time is constructed as
\begin{equation}
  \label{formulation:maxutility}
  U_k(\boldsymbol{\Delta_k}, \boldsymbol{\Delta_{-k}}, R, \theta) = \sum_{t=0}^{T-1} \left(P_k(t) - C_k(t)\right),
\end{equation}

where $\boldsymbol{\Delta_{-k}} = \boldsymbol{\Delta} \backslash \boldsymbol{\Delta_k}$. Given the payment $R$ and conservation rate $\theta$ provided by the server, under the data update dynamics in Equ. (\ref{formulation:update}), the optimization problem for a particular client $k$ can be constructed as
\begin{equation}
  \begin{aligned}
    \max_{\boldsymbol{\Delta_k}=\{\Delta_k(t)\} \atop t \in \{0,\cdots,T-1\}} & U_k(\boldsymbol{\Delta_k}, \boldsymbol{\Delta_{-k}}, R, \theta), \\
    \text{s.t.} & D_k(t + 1) = \theta D_k(t) + \Delta_k(t).
  \end{aligned}  
\end{equation}
Note that a larger $\boldsymbol{\Delta}_k$ enables client $k$ to secure a greater share of the payment, but it also imposes a heavier resource consumption burden. Moreover, according to Equ. (\ref{formulation:Pkt}), client $k$'s optimal strategy $\boldsymbol{\Delta}_k$ is affected by other clients' strategies $\boldsymbol{\Delta_{-k}}$ due to $\sum_{i=1}^N D_i(t)$, which makes the joint data update strategy design over time more challenging.

\subsection{Stackelberg Game Formulation}
\label{section: game}
Based on the discussions in Sections \ref{section:cost function} and \ref{section:utility function}, we formulate the interactions between the server and clients as a two-stage Stackelberg game \cite{tian2024two, wang2023incentive}:
\begin{equation}
  \begin{aligned}
    \mathrm{Stage\ \uppercase\expandafter{\romannumeral1}:} & 
    \min_{R, \theta} U(\boldsymbol{\Delta}, R, \theta), \\
    \mathrm{Stage\ \uppercase\expandafter{\romannumeral2}:} &
    \max_{\boldsymbol{\Delta_k}=\{\Delta_k(t)\} \atop t \in \{0,\cdots,T-1\}} U_k(\boldsymbol{\Delta_k}, \boldsymbol{\Delta_{-k}}, R, \theta), \\
    & \text{s.t.} D_k(t + 1) = \theta D_k(t) + \Delta_k(t).
  \end{aligned}  
\end{equation}

Figure \ref{fig:overview} illustrates the timeline decision process, outlining the sequence of decisions made by both the server and clients in DUFL. At the beginning of the FL tasks, the server launches its optimal strategy $(R, \theta)$ to clients decided in Stage \uppercase\expandafter{\romannumeral1}.
Then in Stage \uppercase\expandafter{\romannumeral2}, according to $(R, \theta)$ announced by the server, each client $k$ decides its optimal strategy $\Delta_k(t)$ to update its buffer. The mutual optimal strategies of the server and clients
form the equilibrium of this Stackelberg game.

\section{Optimal Strategy Analysis}
In this section, we will derive the equilibrium of the Stackelberg game above by backward reduction: first analyze client $k$'s optimal data update strategy $\boldsymbol{\Delta_k}$ in Stage II given any server's strategy $(R, \theta)$, then discuss the optimal payment $R$ and conservation rate $\theta$ of the server in Stage I. 

\subsection{Optimal Strategy of Clients in Stage \uppercase\expandafter{\romannumeral2}}


Before analyzing the data update strategy on clients' side, we face a challenge of incomplete information. Specifically, for a client \( k \), deriving the optimal strategy \( \Delta_k(t) \) at round \( t \) requires knowledge of the total buffered data volume \( \sum_{i=1}^N D_i(t) \) of all clients grounded in Equ.~(\ref{formulation:Pkt}). 
However in reality, \( \sum_{i=1}^{N} D_i(t) \) keeps unknown to client \( k \) due to inter-client information isolation, especially in federated learning. 
To address this, we introduce a mean-field estimator \( \phi(t) \) to approximate \( \sum_{i=1}^{N} D_i(t) \)~\cite{wang2022dynamic, shiri2020communication}. 
Mathematically, \( \phi(t) \) is a given function and regarded as a known term here. 
The estimation of \( \phi(t) \) will be discussed later in Section~\ref{section:estimation}. 

By replacing \( \sum_{i=1}^{N} D_i(t) \) in Equ.~(\ref{formulation:maxutility}) with \( \phi(t) \), the optimization problem of client \( k \) becomes:

\begin{align}
  \label{formulation:reformulated}
  \max_{\boldsymbol{\Delta_k}=\{\Delta_k(t)\} \atop t \in \{0,\cdots,T-1\}} & \sum_{t=0}^{T-1} \left(\frac{D_k(t)}{\phi(t)}R - \alpha_k\Delta_k^2(t) - \beta_kD_k^2(t)\right), \notag \\
  \text{s.t.} & D_k(t+1) = \theta D_k(t) + \Delta_k(t).
\end{align}

Given any strategy $(R, \theta)$ launched by the server, the optimal strategy $\Delta_k(t)$ for each client $k$ at round $t$ is as follows.

\begin{proposition}
  \label{proposition:clientoptimalstrategy}
  For any client $k$ at arbitrary round $t$, the optimal data update strategy $\Delta_k(t)$ is
  \begin{align}
    & \Delta_k(t) = \notag \\
    & \begin{cases}
      \Big[\frac{1}{2 \alpha_k} \sum_{\tau = t + 1}^{T - 1} \theta^{\tau - t - 1}\big(\frac{R}{\phi(\tau)} -&2\beta_k D_k(\tau)\big)\Big]^+, \\ & t \in \{0, \cdots, T-2\}; \\
      0, & t = T-1.
    \end{cases}  
  \label{formulation:delta}
  \end{align}
  And the buffered data volume $D_k(t)$ at round $t$ is
  \begin{align}
    & D_k(t) = \notag \\
    & \begin{cases}
      D_k^0, & t = 0; \\
      \theta^{t} D_k(0) + \sum_{\tau = 0}^{t-1} \theta^{t-1-\tau} \Delta_k(\tau), & t \in \{1, \cdots, T-1\},
    \end{cases}
  \label{formulation:datasize}
  \end{align}
  where $[x]^+ = \max\{x, 0\}, D_k^0$ is the initial buffered data volume of client $k$.
\end{proposition}

The detailed proof is provided in Appendix \ref{proof:clientoptimalstrategy} in the supplementary material.
\begin{remark}
    Equ. (\ref{formulation:delta}) reveals that $\Delta_k(t)$ increases with $R$, which means a greater payment encourages clients to collect more fresh data for buffer updating and model training. In addition, greater collection and training expenditure of $\alpha_k, \beta_k$ inhibit clients from collecting fresh data, which keeps consistent with our intuition.
\end{remark}

\subsection{Optimal Strategy of Server in Stage \uppercase\expandafter{\romannumeral1}}
To achieve the Stackelberg equilibrium, we aim to derive the server's optimal strategy to minimize its cost in response to clients' strategies.
Specifically, provided all clients' optimal strategies of $\boldsymbol{\Delta}$ in Equ. (\ref{formulation:delta}) along with corresponding buffered data volume matrix $\boldsymbol{D}$ and DoS value matrix $\boldsymbol{S}$ where $\boldsymbol{D}=\{D_k(t)\}, \boldsymbol{S}=\{S_k(t)\}, k \in \{1, \cdots, N\}, t \in \{0, \cdots, T-1\}$, we substitute them into the server's objective function in problem (\ref{formulation:mincost}) to derive the optimal strategy of $(R, \theta)$.

However, solving problem (\ref{formulation:mincost}) is non-trivial because it involves two-variable joint optimization including a complicated structure of clients' dynamic strategies as shown in Equ. (\ref{formulation:delta}). 
Therefore, instead of deriving an accurate solution in closed-form, we provide an approximately optimal solution for the server by utilizing Bayesian Optimization (BO), a sample-efficient black-box optimization method which is widely used for low-dimensional expensive-objective problems \cite{badar2024trustfed, zhu2023federated, dai2021differentially}.

In DUFL, BO utilizes a probabilistic surrogate model of Gaussian Process(GP) along with observed data $\mathcal{X} = \{((R_l,\theta_l), U(R_l, \theta_l))\}_{l=1}^p$ to generate a posterior distribution $\hat{U} = \mathbb{P}(U | \mathcal{X})$ over the actual function of $U$ in (\ref{formulation:mincost}). Then the acquisition function of Expected Improvement(EI) leverages $\hat{U}$ to assess the gain of a set of prospective input candidates $\{(R_j, \theta_j)\}_{j=1}^q$ . For any input candidate $(R_j, \theta_j)$, the gain is defined as 
\begin{equation}
EI(R_j,\theta_j) = \mathbb{E}\left[\max(U_{\min} - \hat{U}(R_j,\theta_j), 0)\right],
\end{equation}
where $U_{\min} = \min_l U(R_l, \theta_l)$.
Next, EI selects input candidate with the greatest gain, obtains corresponding output on the actual function of $U$ and adds it into $\mathcal{X}$ for GP model updating. The process continues until convergence, ultimately yielding a near-optimal solution of $(R, \theta)^*$ with minimal server cost.


Note that the convergence of Bayesian Optimization using Gaussian Processes and the Expected Improvement criterion has been theoretically guaranteed under mild regularity conditions in \cite{Srinivas2010}, ensuring reliable iterations toward the global optimum.

\subsection{Algorithm for Finalizing Strategy Design}

\label{section:estimation}
In this paragraph, we aim to find the mean-field estimator $\phi(t)$, thereby finalizing strategy design for this Stackelberg game.
On the one hand, $\phi(t)$ which is constructed to estimate $\sum_{k=0}^{N} D_k(t)$ is affected by the buffered data volume $D_k(t)$ of all clients, and further affected by their data update strategies $\Delta_k(\tau), \tau \in \{0,\cdots,t-1\}$ according to Equ. (\ref{formulation:datasize}); 
on the other hand, $\phi(t)$ will in turn affect the determination of $\Delta_k(\tau), \tau \in \{0,\cdots,t-1\}$ according to Equ. (\ref{formulation:delta}).
In brief, there is a close loop among $\phi(t), D_k(t)$ and $\Delta_k(\tau), \tau \in \{0, \cdots, t-1\}$. Based on this, we have the following proposition.
\begin{proposition}
  \label{proposition:fixed point}
  There exists a fix point for the mean-field estimator $\phi(t), t \in \{0, \cdots, T - 1\}$.
\end{proposition}
The detailed proof is provided in Appendix \ref{proof:fixed_point} in the supplementary material.

Based on Proposition \ref{proposition:fixed point}, we develop a fixed-point method to solve $\phi(t)$ and the completed algorithm for DUFL is summarized in Algorithm \ref{alg:DUFL}:
In the \textit{Strategy Decision Phase}, the server first initializes mean-field estimator $\phi_0(t)$.
Then, fixed-point iterations are performed where the server optimizes $(R, \theta)$ to minimize its own cost function via Bayesian Optimization, while clients compute their prospective data increments $\Delta_k(t)$ and corresponding buffer volumes $D_k(t)$.
The mean-field estimator $\phi(t)$ is updated accordingly until convergence. Once the mean-field estimator $\phi(t)$ converges, both the server's strategy $(R, \theta)$ and clients' local strategies $\mathbf{\Delta}$ converge synchronously.

Afterwards, the \textit{Federated Training Phase} begins.
In each communication round, the server broadcasts the global model to clients.
Clients update their local buffers by discarding a fraction of stale data and incorporating new samples according to the finalized strategy, then perform local model updates based on their current buffers.
The updated models are uploaded and aggregated at the server to form the next global model.
This training process continues until the final round is reached.

The overall time complexity of DUFL is composed of two phases. In the strategy finalization phase, Bayesian optimization incurs $O(M^4)$ complexity for Gaussian process training and $O(M^2 {d_i}^3)$ complexity for acquisition optimization, where $M$ is the number of Bayesian optimization steps and $d_i$ is the dimension of the input space. Then client-side computing for data update adds $O(N T^2)$ per iteration, with $N$ denoting the number of clients and $T$ the total number of communication rounds. Provided that the fixed-point iteration needs to be repeated for $J$ times to achieve convergence, this phase requires $O\left( J (M^4 + M^2 {d_i}^3 + N T^2) \right)$. 
In the federated training phase, clients update models with $d_m$ number of model parameters over $E$ epochs, leading to a complexity of $O\left( T N E {d_m}^2 \right)$.
Therefore, the total time complexity is $O\left( J (M^4 + M^2 {d_i}^3 + N T^2) + T N E {d_m}^2 \right).$
\begin{remark}
While the theoretical complexity includes polynomial terms in $M$ and $d_i$, we note that in practice, these quantities are kept small (e.g., $J=20$, $M=30$, $d_i=2$) due to rapid convergence rate of both fixed-point method and Bayesian Optimization in practice, which leads to managable computational cost in all our experiments.
\end{remark}

\begin{algorithm}[H]
    \caption{DUFL Mechanism}
    \label{alg:DUFL}
    \begin{algorithmic}[1]
        \STATE{\bfseries Input:} Number of rounds $T$, number of clients $N$, initial buffer volume $D_k(0)$, hyperparameters.
        \STATE{\bfseries Output:} Final global model $w(T)$.

        \vspace{0.5em}
        \STATE{\textbf{\underline{Strategy Decision Phase}}}
        \vspace{0.3em}

        \STATE Initialize mean-field estimator $\phi_0(t)$ for all $t = 0,\dots,T-1$;  iteration counter $j=0$.
        \REPEAT
            \STATE Given $\phi_j(t)$, server optimizes $(R, \theta)$ via Bayesian Optimization.
            \FOR{each client $k = 1$ to $N$}
                \FOR{each round $t = 0$ to $T-1$}
                    \STATE Calculate data increment: $\Delta_k(t) = \left[\frac{1}{2\alpha_k} \sum_{\tau=t+1}^{T-1} \theta^{\tau-t-1} \left( \frac{R}{\phi(\tau)} - 2\beta_k D_k(\tau) \right) \right]^+$.
                    \STATE Calculate buffer volume: $D_k(t+1) = \theta D_k(t) + \Delta_k(t)$.
                \ENDFOR
            \ENDFOR
            \STATE Update mean-field estimator: $\phi_{j+1}(t) = \sum_{k=1}^{N} D_k(t)$ for all $t$.
            \STATE $j \gets j + 1$.
        \UNTIL{$\max_t |\phi_j(t) - \phi_{j-1}(t)| \leq \epsilon$}.

        \vspace{0.5em}
        \STATE{\textbf{\underline{Federated Training Phase}}}
        \vspace{0.3em}

        \FOR{each round $t = 0$ to $T-1$}
            \STATE Server broadcasts the global model $w(t)$ to all clients.
            \FOR{each client $k$ in parallel}
                \STATE Discard $(1-\theta)D_k(t-1)$ stale data points from buffer.
                \STATE Collect $\Delta_k(t-1)$ new data points into buffer.
                \STATE Perform local model update: $w_k(t+1) = w(t) - \eta \nabla F_k(w(t) \mid D_k(t))$.
                \STATE Upload updated local model $w_k(t+1)$ to server.
            \ENDFOR
            \STATE Server aggregates models: $w(t+1) = \sum_{k=1}^{N} \frac{D_k(t)}{D(t)} w_k(t+1)$.
        \ENDFOR
    \end{algorithmic}
\end{algorithm}

\section{Experiment}
In this section, we evaluate the performance of proposed DUFL by numerical experiments. Note that due to page limit, additional experiments are provided in Appendix B in the supplementary material.
\subsection{Experimental Setup}
\textbf{Datasets and Models: }


Our experiments are conducted on five widely used benchmarks: MNIST, FMNIST, SVHN, CIFAR-10, and CIFAR-100. We employ LR for MNIST and FMNIST, resulting in convex optimization problems aligned with our theoretical assumptions. For SVHN and CIFAR-10, we use a lightweight CNN, and for CIFAR-100, ResNet-18, to validate our method’s robustness in non-convex scenarios.

To simulate time-sensitive federated learning, two dynamics are introduced. First, to model data aging, client-side samples are progressively blurred via additive noise over communication rounds. Second, to emulate evolving label distributions, each client starts with only a subset of classes, with new classes introduced as training proceeds. Both blurring severity and initial class coverage are governed by a time sensitivity coefficient $\sigma$, where larger values imply stronger noise and fewer initial classes.

\textbf{Baselines: }
We validate our proposed DUFL mechanism on several classical federated learning baseline algorithms, including (1)FedAvg \cite{mcmahan2017communication}, (2)FedProx \cite{li2020federated}, (3)FedDyn \cite{acar2021federated}. 


\textbf{Hyperparameters Settings: }
We simulate $T=100$ federated rounds, each involving $N=15$ clients by default. Each client initializes with $D_k^0=1000$ samples, and adopts SGD for 20 local epochs using batch size 64 and learning rate $\eta=10^{-2}$. Unit costs of data collection and training follow $\alpha_k \sim \mathcal{U}(10^{-4}, 10^{-3})$ and $\beta_k \sim \mathcal{U}(5\times10^{-6}, 5\times10^{-5})$. The server sets trade-off factor $\gamma=10^{-4}$ and uses $\kappa_1 = \kappa_2 = 1$, $\kappa_3 = 10^{-2}$ in its utility. The strategy space includes $\theta \in [0, 1]$ and payment level $R \in [0, 500]$. Default time sensitivity coefficient is $\sigma = 0.75$.

\subsection{Utility Evaluation}
We provide utility evaluation from the prospective of both the server and clients in this section.

\begin{figure}[htbp]
  \centering
  \begin{subfigure}{0.465\columnwidth}
    \centering
    \includegraphics[width=\linewidth]{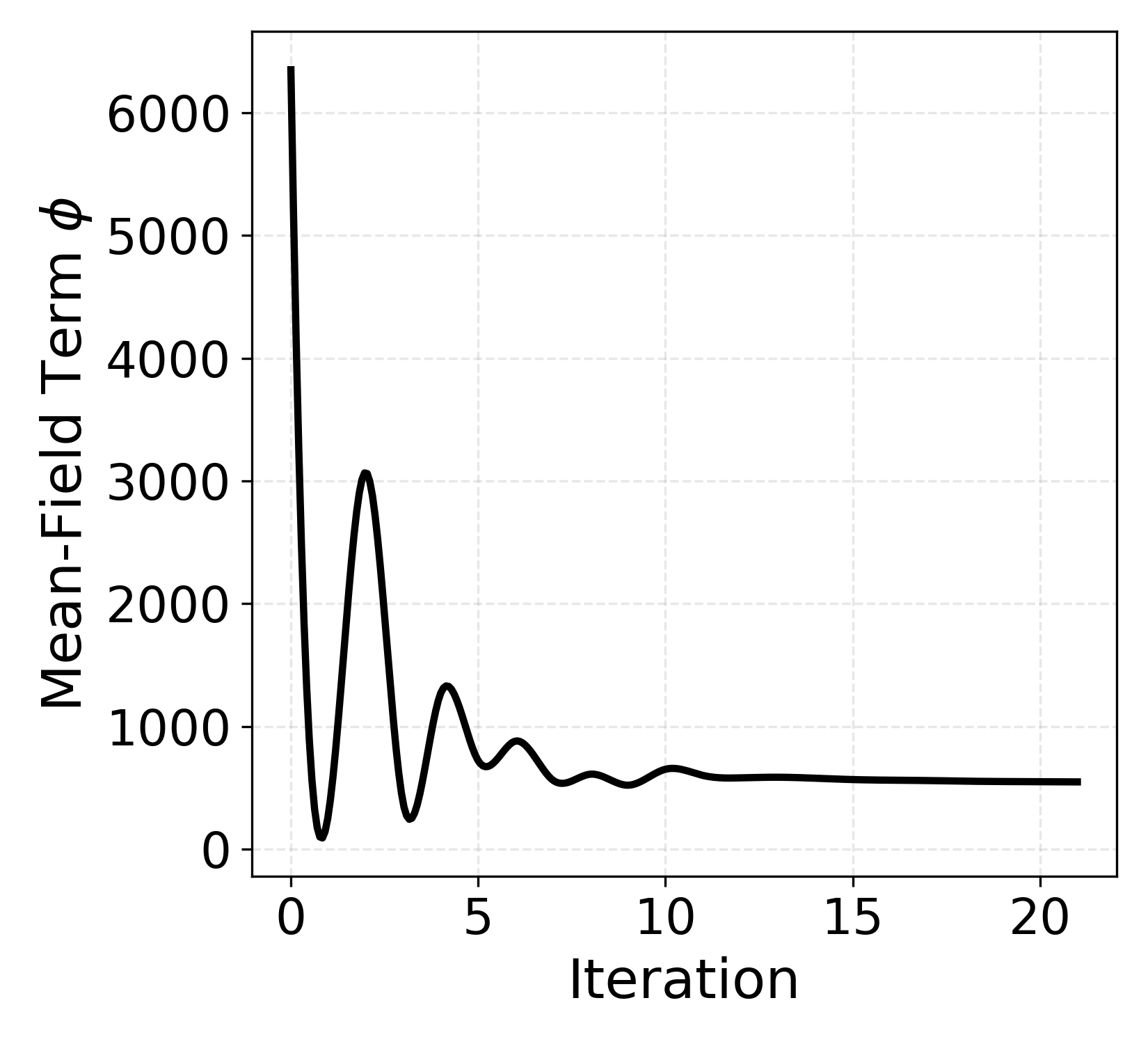}
  \end{subfigure}
  \begin{subfigure}{0.520\columnwidth}
    \centering
    \includegraphics[width=\linewidth]{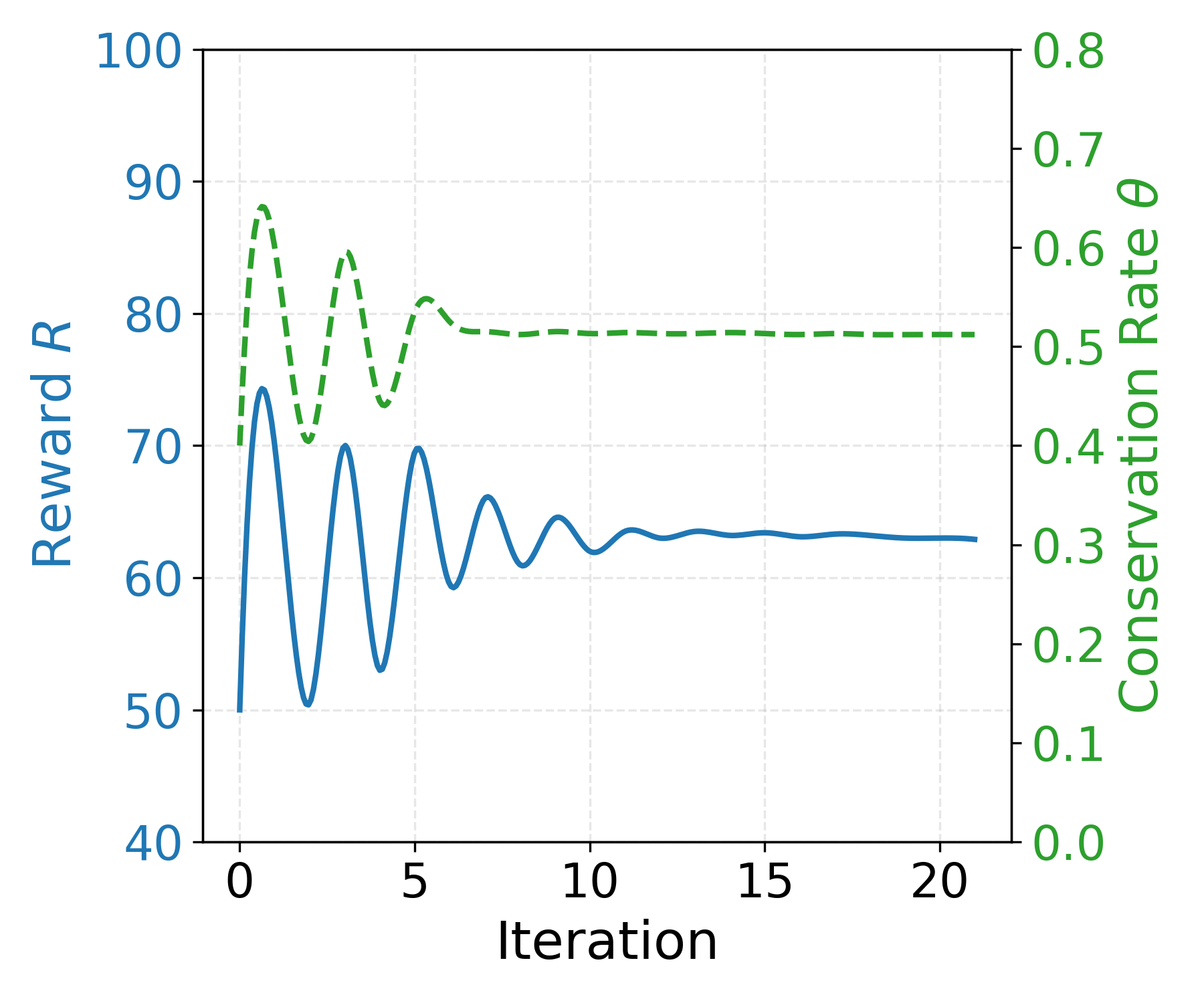}
  \end{subfigure}
  \caption{Illustration of convergence behavior for mean-field estimator $\phi$(\textbf{Left}) and server's strategy $(R, \theta)$(\textbf{Right}).}
  \vspace{1.5em}
  \label{fig:convergence}
\end{figure}

\textbf{Fixed-Point Convergence Process:}
The iteration process of fixed-point convergence is depicted in Figure \ref{fig:convergence}. We find that mean-field estimator $\phi$ and the server's strategy $(R, \theta)$ converge synchronously 
 and rapidly in a few steps, which substantiates the efficiency of our proposed DUFL algorithm.

\textbf{Effect of Client's Strategy on the Utility: }
In this paragraph, we analyze the effect of data update strategy $\Delta_k(t)$ on clients' utility and the results are plotted in Figure \ref{fig:clients}.
For comparison, we implement two auxiliary strategies: zero strategy and random strategy. For a particular client $k$, zero strategy means it doesn't collect any new data through the training procedure, while random strategy refers that it collects new data randomly in each time slot. Note that in order to keep comparison meaningful, the total amount of new data collected over the time horizon under random strategy is set to be roughly consistent with that under optimal strategy.
It is shown that the optimal strategy helps client $k$ secure the highest utility compared with other two strategies.
Provided that clients are selfish, this indicates that each of them will follow the optimal strategy, thereby the mutually best response strategies are reached simultaneously, which meets the requirement of the Nash equilibrium solution of Stage \uppercase\expandafter{\romannumeral2}.
In addition, we can find under the same data update strategy, the client's utility $U_k$ decreases with number of clients $N$. The underlying reason is that client expansion may intensify the competition for the reward, thereby leading to reward reduction and utility reduction.

\textbf{Effect of Server's Strategy on the Cost: }
In this paragraph, we analyze the effect of server's strategy $(R, \theta)$ on its cost. Under the hyperparameter settings in experimental setup, the server's optimal strategy is derived as $(R, \theta)^* = (63.18, 0.52)$.
For comparison, we fix the optimal payment $R^*$ and consider three types of strategies: $(R, \theta)^1 = (63.18, 0.12), (R, \theta)^2 = (63.18, 0.82)$ and $(R, \theta)^3 = (63.18, 0.92)$.
Besides, we also fix the optimal conservation rate $\theta^*$ and consider another three types of strategies: $(R,\theta)^4 = (13.18, 0.52), (R, \theta)^5 = (203.18, 0.52)$ and $(R, \theta)^6 = (423.18, 0.52)$.

\noindent
\begin{minipage}[t]{0.49\linewidth}
  \centering
  \includegraphics[width=\linewidth]{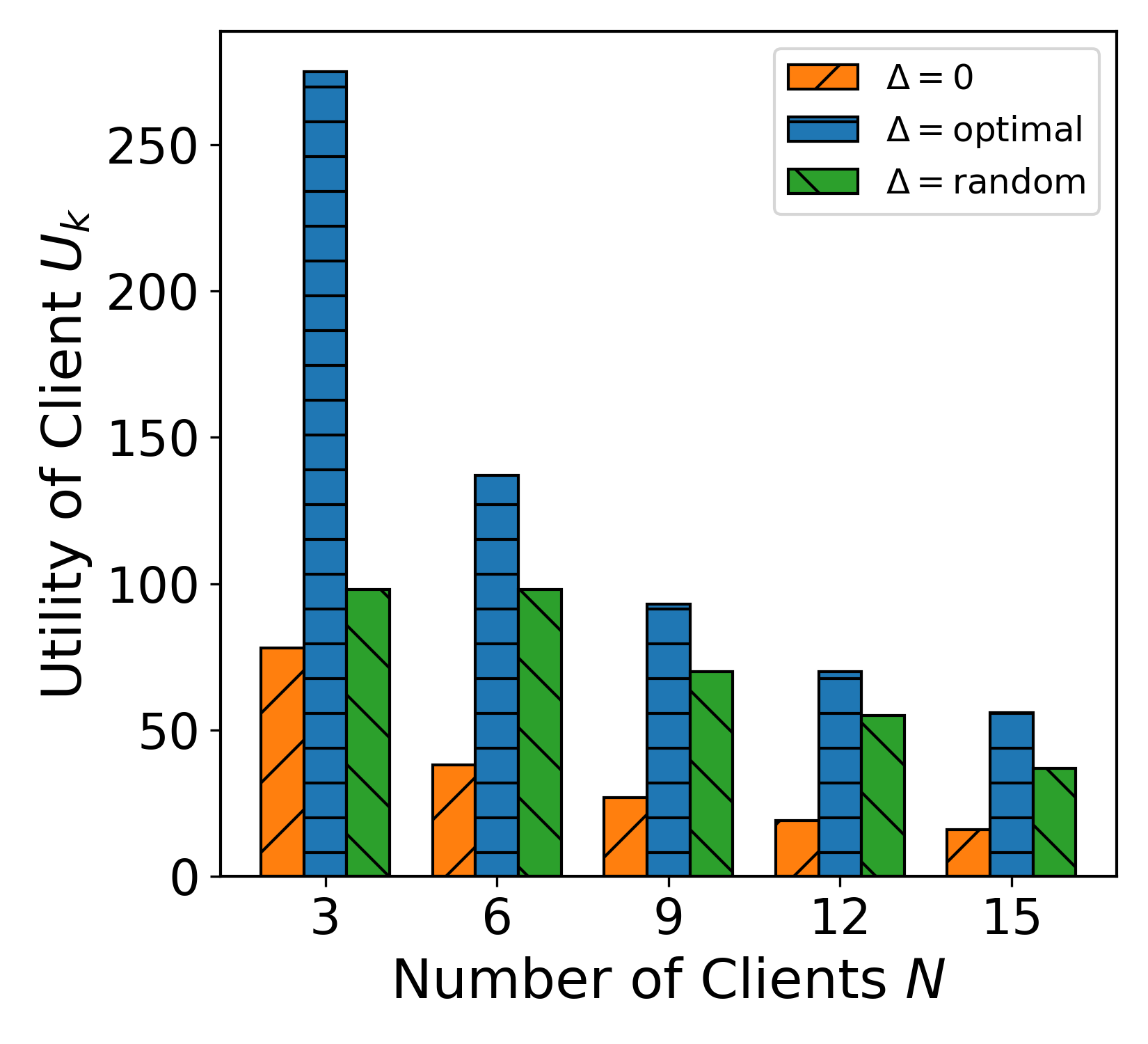}
  \captionof{figure}{Comparative analysis of utility function $U_k$ of client $k$ over various strategies $\mathbf \Delta_k$.}
  \label{fig:clients}
  \vspace{1.5em}
\end{minipage}%
\hfill
\begin{minipage}[t]{0.49\linewidth}
  \centering
  \includegraphics[width=\linewidth]{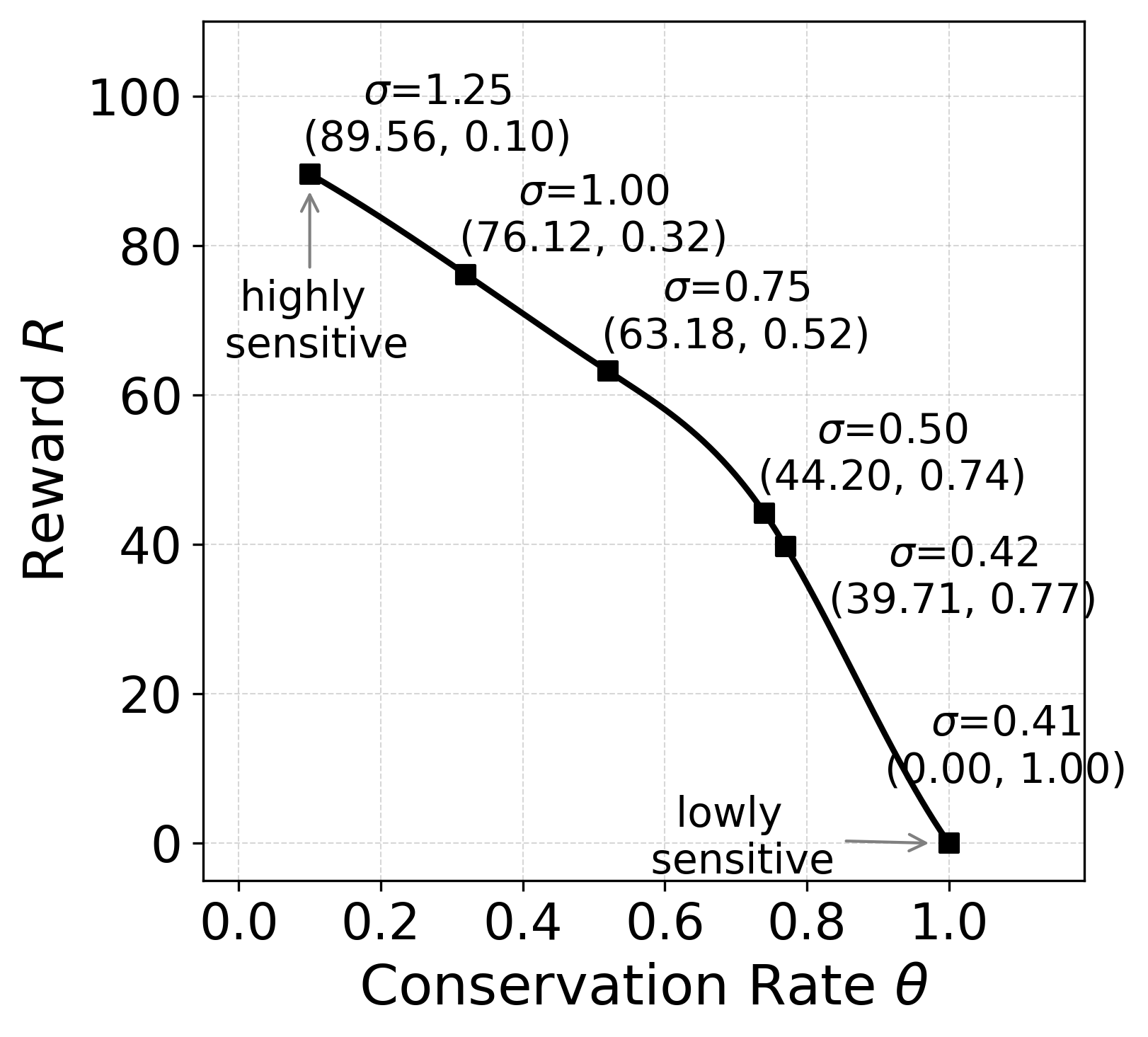}
  \captionof{figure}{The effect of time sensitivity coefficient $\sigma$ on the server's determination of strategy $(R, \theta)$.}
  \label{fig:hyperparam}
  \vspace{1.5em}
\end{minipage}

\begin{figure}[b]
    \centering
    \includegraphics[width=\columnwidth]{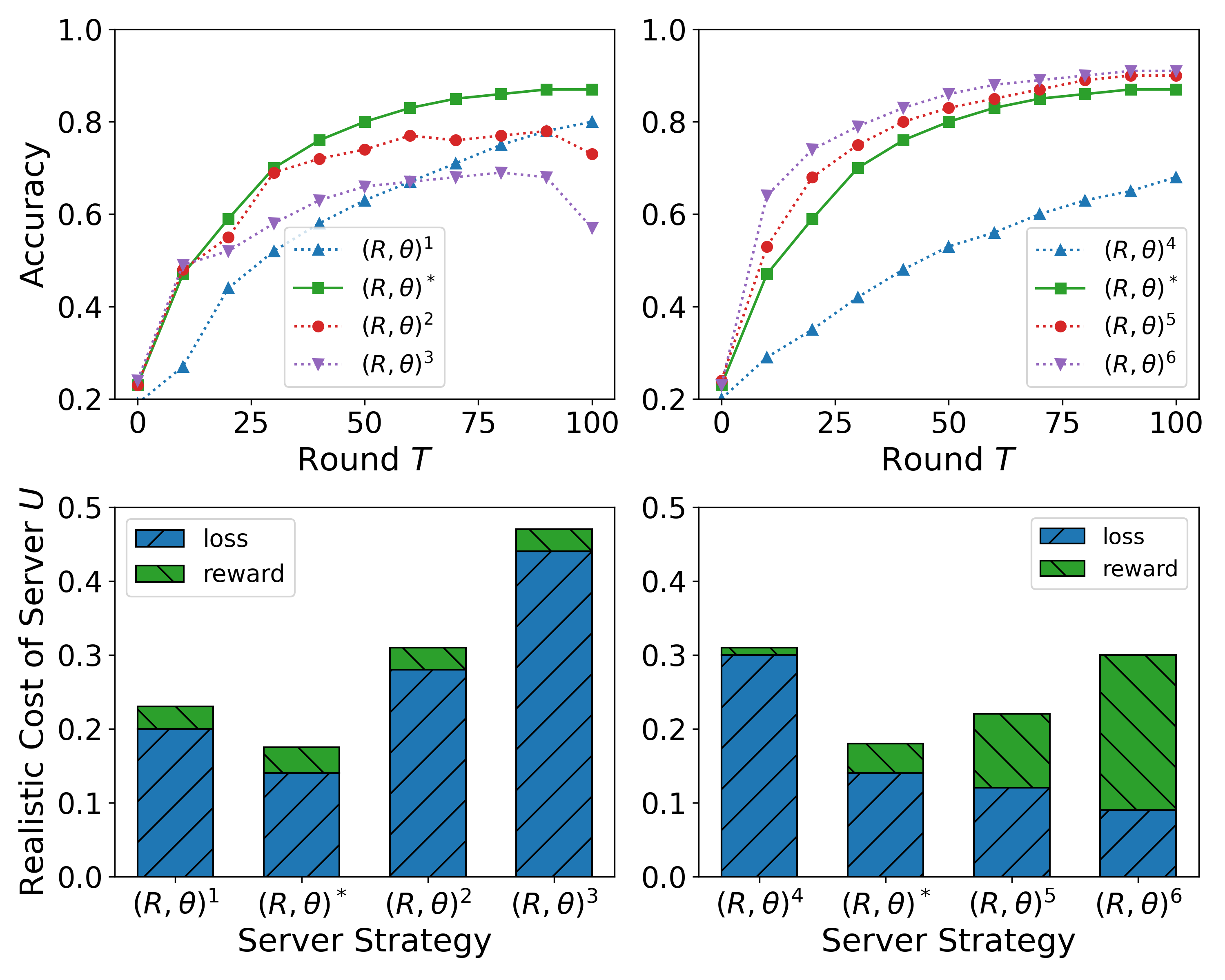}
    \caption{Comparative analysis of test accuracy (\textbf{Top}) and cost (\textbf{Bottom}) versus communication rounds $T$ under various strategies with CNN over Cifar-10.}
    \label{fig:server_cost_mnist}
    \vspace{1.5em}
\end{figure}

The effect of $\theta$ and $R$ on test accuracy and cost is shown in Figures \ref{fig:server_cost_mnist}.
We can find that $(R,\theta)^*$(solid line) outperforms $(R,\theta)^1, (R,\theta)^2$ and $(R,\theta)^3$(dotted line) in terms of test accuracy on the Cifar-10 datasets. The underlying reason is that strategy $(R,\theta)^2, (R,\theta)^3$ with improperly great $\theta$ contributes to huge but stale buffered data, while $(R,\theta)^1$ with improperly small $\theta$ contributes to fresh but small amount of data. Both of them lead to worse model performance.
Under the same payment to clients, $(R,\theta)^*$ reaches the lowest cost compared with other strategies.

In addition, $(R,\theta)^5$ and $(R,\theta)^6$(dotted line) surpass the optimal strategy of $(R,\theta)^*$(solid line) with respect to test accuracy across all datasets because richer payment encourages clients to collect more fresh data, thereby enhance the model performance.
However, $(R,\theta)^*$ still reach the lowest cost, which indicates that our proposed strategy has the ability to strike the trade-off between payment and accuracy loss simultaneously.

\textbf{Effect of time sensitivity coefficient on Server's Strategy: }
In this paragraph, we explore the effect of time sensitivity coefficient $\sigma$ on the server's strategy $(R, \theta)$.
Specifically, we set $\sigma$ to be $0.41, 0.42, 0.50, 0.75, 1.00, 1.25$, where larger $\sigma$ indicates stronger sensitivity to data staleness.

It is shown in Figure \ref{fig:hyperparam} that, as $\sigma$ increases, the optimal reward $R$ rises while the optimal conservation rate $\theta$ drops.
It matches the intuition that when facing tasks with higher time sensitivity, the value of previous data decreases. The server needs to guide clients timely discarding outdated data and collecting fresh data. Note that data update ceases when $\sigma = 0.41$ or below, indicating that the server tolerates mild staleness when the expected accuracy gain no longer offsets the update cost.
The result demonstrates that our proposed strategy is adaptable to federated learning tasks with varying time sensitivity. 
\subsection{Performance Evaluation}
\textbf{Comparison of Model Performance: }
In this paragraph, we analyze model performance of our proposed DUFL compared with 3 baseline algorithms over 5 federated learning tasks under various $\sigma$.
Specifically, all baseline algorithms are evaluated under two configurations. In the first setting, models are trained directly on the initial static local datasets without any data update throughout the training process, corresponding to the standard practice in traditional federated learning. In the second setting, our proposed data update mechanism is integrated as a plugin, enabling clients to discard outdated samples and incorporate newly collected data during training. 

The results are summarized in Table \ref{table:acc}. In highly time-sensitive tasks($\sigma=1.25$) with severe data aging and distribution drift, our proposed DUFL framework achieves tremendous accuracy improvement across all datasets compared with baseline algorithms. It validates that DUFL can effectively mitigate the adverse effects of data staleness via positive data update.
In lowly time-sensitive tasks($\sigma=0.00$) where data value remains relatively stable, DUFL performs closely to baseline algorithms. The reason is that under this circumstance, present data stay valuable to train model and it's not necessary for DUFL to update data considering monetary cost, which is essentially equivalent to baseline algorithms.

\begin{table}[htbp]
\centering
\caption{Performance comparison across different methods and datasets under varying time sensitivity coefficient $\sigma$.}
\resizebox{\columnwidth}{!}{%
\begin{tabular}{cc | cc | cc | cc}
\toprule
\textbf{Dataset} & $\boldsymbol{\sigma}$ & \multicolumn{2}{c}{\textbf{FedAvg}} & \multicolumn{2}{c}{\textbf{FedProx}} & \multicolumn{2}{c}{\textbf{FedDyn}}\\
\textbf{+Model}  &                       & Std. & +DUFL & Std. & +DUFL & Std. & +DUFL\\
\midrule
MNIST & 1.25 & 57.9\% & 91.8\% & 58.1\% & 91.7\% & 58.8\% & 91.5\% \\
+LR   & 0.75 & 76.2\% & 91.2\% & 76.9\% & 92.4\% & 77.1\% & 92.0\% \\
(convex) & 0.00 & 91.6\% & 91.6\% & 92.7\% & 92.7\% & 92.7\% & 92.7\% \\
\midrule
FMNIST & 1.25 & 53.8\% & 79.1\% & 54.2\% & 79.8\% & 55.0\% & 80.7\% \\
+LR    & 0.75 & 66.6\% & 80.8\% & 67.4\% & 81.8\% & 67.1\% & 82.6\% \\
(convex) & 0.00 & 84.3\% & 84.3\% & 85.1\% & 85.1\% & 84.9\% & 84.9\% \\
\midrule
SVHN & 1.25 & 69.4\% & 91.3\% & 70.2\% & 92.0\% & 72.4\% & 93.5\% \\
+CNN & 0.75 & 83.6\% & 92.7\% & 83.9\% & 93.1\% & 85.0\% & 93.7\% \\
(non-convex) & 0.00 & 90.9\% & 90.9\% & 92.7\% & 92.7\% & 93.1\% & 93.1\% \\
\midrule
CIFAR-10 & 1.25 & 54.3\% & 82.4\% & 55.7\% & 82.6\% & 58.5\% & 84.4\% \\
+CNN     & 0.75 & 72.2\% & 83.9\% & 72.3\% & 85.5\% & 74.1\% & 83.9\% \\
(non-convex) & 0.00 & 81.5\% & 81.5\% & 84.3\% & 84.3\% & 84.7\% & 84.7\% \\
\midrule
CIFAR-100 & 1.25 & 44.5\% & 56.7\% & 46.2\% & 56.9\% & 47.8\% & 59.0\% \\
+ResNet-18 & 0.75 & 54.9\% & 57.1\% & 56.8\% & 58.0\% & 59.2\% & 60.1\% \\
(non-convex) & 0.00 & 57.6\% & 57.6\% & 59.2\% & 59.2\% & 63.5\% & 63.5\% \\
\bottomrule
\end{tabular}
}
\label{table:acc}
\end{table}

\section{Conclusion}
In this paper, we propose an innovative framework DUFL for federated learning with highly time-sensitive tasks, where a new dynamic data update scheme is designed to balance data staleness and data volume simultaneously. A novel DoS metric is introduced, based on which theoretical analyses are conducted. We model the interactions of the server and clients as a Stackelberg game and derive the optimal strategies for both the server and clients. Extensive experiments demonstrate the superiority of our proposed approach. 




\bibliography{m6810}

\clearpage
\setcounter{page}{1} 
\appendix
In the appendix, the complete proofs of theoretic results provided in the main text and additional experiments are exhibited in detail.
\section{Proof of Theoretic Results}
\subsection{Proof of Lemma \ref{lemma:generalformula}}
\label{proof:generalformula}
\begin{proof}
    According to the definition of DoS, we have
    \begin{align}
        & S_k(t+1) \\
        = & S_k(t) \frac{\theta D_k(t)}{D_k(t+1)} + 1 \notag \\
        = & S_k(t)\left(1 - \frac{\Delta_k(t)}{D_k(t+1)}\right) + 1 \notag \\
        = & \left(S_k(t-1) \frac{\theta D_k(t-1)}{D_k(t)} + 1\right) \frac{\theta D_k(t)}{D_k(t+1)} + 1 \notag \\
        = & S_k(t-1) \frac{\theta^2 D_k(t-1)}{D_k(t+1)} + \frac{\theta D_k(t)}{D_k(t+1)} + 1 \notag \\
        = & S_k(t-2) \frac{\theta^3 D_k(t-2)}{D_k(t+1)} + \frac{\theta^2 D_k(t-1)}{D_k(t+1)} + \frac{\theta D_k(t)}{D_k(t+1)} + 1 \notag \\
        = & \cdots \notag \\
        = & S_k(0) \frac{\theta^{t+1} D_k(0)}{D_k(t+1)} + \frac{\theta^t D_k(1)}{D_k(t+1)} + \cdots + \frac{\theta^2 D_k(t-1)}{D_k(t+1)} \notag \\
          & + \frac{\theta D_k(t)}{D_k(t+1)} + 1 \notag \\
        = & \sum_{\tau = 0}^{t+1} \frac{\theta^{t+1-\tau} D_k(\tau)}{D_k(t+1)}.
    \end{align}
  \end{proof}

\subsection{Proof of Theorem \ref{theorem:upperbound}}
\begin{proof}
  \label{proof:upperbound}
  Let's pay attention to the round $t+1$.
  \begin{align}
    \label{proof:init}
      & \mathbb E[F(w(t+1)|\mathcal D(t+1))-F(w(t)|\mathcal D(t))] \notag \\
      = & \mathbb E[F(w(t+1)|\mathcal D(t))-F(w(t)|\mathcal D(t))] \notag \\
        & + \mathbb E[F(w(t+1)|\mathcal D(t+1))-F(w(t+1)|\mathcal D(t))] \notag \\
      = & \underbrace{\mathbb E[F(w(t+1)|\mathcal D(t))-F(w(t)|\mathcal D(t))]}_A + \Omega_t,
  \end{align}
  where $\Omega_t = E[F(w(t+1)|\mathcal D(t+1))-F(w(t+1)|\mathcal D(t))]$. It captures the expected difference of global loss function based on the same global model $w(t)$ between buffered data at present round $\mathcal D(t)$ and at previous round $\mathcal D(t - 1)$. 
  Then we focus on $A$. Since $F_k(w)$ is $\beta$-Lipschitz smooth, we have
  \begin{align}
    \label{proof:boundA}
         & \mathbb E[F(w(t+1)|\mathcal D(t))-F(w(t)|\mathcal D(t))] \notag \\
    \leq & \mathbb E\left\langle \nabla F(w(t)|\mathcal D(t)), w(t+1)-w(t)|\mathcal D(t)\right\rangle \notag \\ 
         & + \frac{\beta}{2}E{\Vert w(t+1)-w(t)|\mathcal D(t) \Vert}^2 \notag \\
    =    & \mathbb E\left\langle \nabla F(w(t)|\mathcal D(t)), (-\eta)\nabla F(w(t)|\mathcal D(t))\right\rangle \notag \\
         & + \frac{\beta}{2}\mathbb E{\Vert w(t+1)-w(t)|\mathcal D(t) \Vert}^2 \notag \\ 
    =    & (-\eta) \Vert \nabla F(w(t)|\mathcal D(t))\Vert^2 \notag \\
         & + \underbrace{\frac{\beta}{2}\mathbb E{\Vert w(t+1)-w(t)|\mathcal D(t) \Vert}^2}_{B}.
  \end{align}
  Next, we focus on bounding $B$. Because the stochastic gradient of $\nabla F_k(w)$ is unbiased and variance-bounded, we have 
  \begin{align}
    \label{proof:boundB}
         & \frac{\beta}{2} \mathbb E {\Vert w(t+1) - w(t) | \mathcal D(t) \Vert}^2 \notag \\
    =    & \frac{\beta}{2} \mathbb E {\Vert (-\eta) \nabla F(w(t)|\mathcal D(t))\Vert}^2 \notag \\
    \leq & \beta\eta^2 \mathbb E{\Vert\nabla F(w(t)|\mathcal D(t))\Vert}^2 \notag \\
    =    & \beta\eta^2 \mathbb E{\left\Vert \sum_{k=1}^{N} \frac{D_k(t)}{D(t)} \nabla F_k(w(t)|\mathcal D_k(t))\right\Vert}^2 \notag \\
    \leq & \beta\eta^2 \sum_{k=1}^{N} \frac{D_k(t)}{D(t)} \mathbb E\Vert\nabla F_k(w(t)|\mathcal D_k(t))\Vert^2 \notag \\
    \leq & \beta\eta^2 \sum_{k=1}^{N} \frac{D_k(t)}{D(t)} \left(2\Vert\nabla F(w(t)|\mathcal D(t))\Vert^2 + \frac{\psi^2}{D_k(t)} + S_k(t)\sigma^2\right) \notag \\
    =    & 2\beta\eta^2 \Vert\nabla F(w(t)|\mathcal D(t))\Vert^2 + \beta\eta^2 \frac{N\psi^2}{D(t)} + \beta\eta^2 \sum_{k=1}^{N} \frac{D_k(t)}{D(t)} S_k(t) \sigma^2.
  \end{align}
  Substituting Equ. (\ref{proof:boundB}) into Equ. (\ref{proof:boundA}), we have
  \begin{align}
    \label{proof:substituteB}
    & \mathbb E[F(w(t+1)|\mathcal D(t)) - F(w(t)|\mathcal D(t))] \notag \\
    \leq & \underbrace{ (2\beta \eta^2 - \eta) \mathbb E{\Vert\nabla F(w(t)|\mathcal D(t))\Vert}^2}_{C} + 2\beta\eta^2 \frac{N\psi^2}{D(t)} \notag \\
    + & \beta\eta^2 \sum_{k=1}^{N} \frac{D_k(t)}{D(t)} S_k(t) \sigma^2.
  \end{align}
  Now we bound $C$. We set $\eta < \frac{1}{2\beta}$, then $2\beta\eta^2 - \eta < 0$. Polyak Lojasiewicz condition holds for $F_k(w)$ due to strong convexity in assumptions:
  \begin{equation}
    \mathbb E[F(w(t)|\mathcal D(t))-F(w^*)] \leq \frac{1}{2\mu}\mathbb E{\Vert\nabla F(w(t)|\mathcal D(t))\Vert}_2^2.
  \end{equation}
  Thus the following inequality holds:
  \begin{align}
    \label{proof:boundC}
         & 2\mu (2\beta\eta^2 - \eta)\mathbb E[(F(w(t)|\mathcal D(t))-F(w^*))] \notag \\
    \geq & (2\beta \eta^2 - \eta) \mathbb E{\Vert\nabla F(w(t)|\mathcal D(t))\Vert}_2^2.
  \end{align}
  Substituting Equ. (\ref{proof:boundC}) into Equ. (\ref{proof:substituteB}), we have
  \begin{align}
    \label{proof:substituteC}
         & \mathbb E[F(w(t+1)|\mathcal D(t)) - F(w(t)|\mathcal D(t))] \notag \\
    \leq & 2 \mu(2\beta\eta^2-\eta)\mathbb E[F(w(t)|\mathcal D(t))-F(w^*)] \notag \\
    +    & 2\beta\eta^2 \frac{N\psi^2}{D(t)} + \beta\eta^2 \sum_{k=1}^{N} \frac{D_k(t)}{D(t)} S_k(t) \sigma^2.
  \end{align}
  Consequently, substituting Equ. (\ref{proof:substituteC}) into Equ. (\ref{proof:init}), we have
  \begin{align}
         & \mathbb E[F(w(t+1)|\mathcal D(t+1))-F(w(t)|\mathcal D(t))] \notag \\
    =    & \mathbb E[F(w(t+1)|\mathcal D(t))-F(w(t)|\mathcal D(t))] + \Omega_t \notag \\
    \leq & 2 \mu(2\beta\eta^2-\eta)\mathbb E[F(w(t-1)|\mathcal D(t))-F(w^*|\mathcal D(t))] \notag \\
    +    & 2\beta\eta^2 \frac{N\psi^2}{D(t)} + \beta\eta^2 \sum_{k=1}^{N} \frac{D_k(t)}{D(t)} S_k(t) \sigma^2 + \Omega_t.
  \end{align}
  Adding $\mathbb E[F(w(t)|\mathcal D(t))-F(w^*)]$ on both sides, we get
  \begin{align}
    \label{proof:preend}
         & \mathbb E[F(w(t+1)|\mathcal D(t+1)) - F(w^*)] \notag \\
    \leq & (1+4\mu\beta\eta^2-2\mu\eta)\mathbb E[F(w(t)|\mathcal D(t))-F(w^*)] \notag \\
    +    & 2\beta\eta^2 \frac{N\psi^2}{D(t)} + \beta\eta^2 \sum_{k=1}^{N} \frac{D_k(t)}{D(t)} S_k(t) \sigma^2 + \Omega_t.
  \end{align}
  Then call Equ. (\ref{proof:preend}) recursively, we have
  \begin{align}
         & \mathbb E[F(w(t+1)|\mathcal D(t+1)) - F(w^*)] \notag \\
    \leq & (1+4\mu\beta\eta^2-2\mu\eta)\mathbb E[F(w(t)|\mathcal D(t))-F(w^*)] \notag \\
         & + 2\beta\eta^2 \frac{N\psi^2}{D(t)} + \beta\eta^2 \sum_{k=1}^{N} \frac{D_k(t)}{D(t)} S_k(t) \sigma^2 + \Omega_t \notag \\
    \leq & (1+4\mu\beta\eta^2-2\mu\eta)^2 \mathbb E[F(w(t-1)|\mathcal D(t-1))-F(w^*)] \notag \\
         & + (1+4\mu\beta\eta^2-2\mu\eta) \times \notag \\
         & \left[2\beta\eta^2 \frac{N\psi^2}{D(t-1)} + \beta\eta^2 \sum_{k=1}^{N} \frac{D_k(t-1)}{D(t-1)} S_k(t-1) \sigma^2 + \Omega_{t-1}\right] \notag \\
         & + \left[2\beta\eta^2 \frac{N\psi^2}{D(t)} + \beta\eta^2 \sum_{k=1}^{N} \frac{D_k(t)}{D(t)} S_k(t) \sigma^2 + \Omega_t \right] \notag \\
    \leq & \cdots \notag \\
    \leq & (1+4\mu\beta\eta^2-2\mu\eta)^{t+1}\mathbb E[F(w(0)|\mathcal D(0))-F(w^*)] \notag \\
         & + \sum_{r=0}^t(1+4\mu\beta\eta^2-2\mu\eta)^r \times \notag \\
         & \left[2\beta\eta^2 \frac{N\psi^2}{D(t-r)} + \beta\eta^2 \sum_{k=1}^{N} \frac{D_k(t-r)}{D(t-r)} S_k(t-r) \sigma^2 + \Omega_{t-r} \right].
  \end{align}
  For ease of representation, let $\kappa_1 = 1 + 4\mu\beta\eta^2 - 2\mu\eta$, $\kappa_2 = 2\beta\eta^2$ and $\kappa_3 = \beta\eta^2$.
  Thus, the convergence upper bound after $T+1$ rounds can be reformulated as
  \begin{align}
         & \mathbb E[F(w(T+1)|\mathcal D(T+1)) - F(w^*)] \notag \\
    \leq & \kappa_1^{T+1} \mathbb E[F(w(0)|\mathcal D(0))-F(w^*)] \notag \\
         & + \sum_{t=0}^T\kappa_1^t \left[\kappa_2 \frac{N\psi^2}{D(T-t)} + \kappa_3 \sum_{k=1}^{N} \frac{D_k(T-t)}{D(T-t)} S_k(T-t) \sigma^2 + \Omega_{T-t} \right] \notag \\
    =    & \kappa_1^{T+1} \mathbb E[F(w(0)|\mathcal D(0))-F(w^*)] \notag \\
         & + \sum_{t=0}^{T}\kappa_1^{T-t} \left[\kappa_2 \frac{N\psi^2}{D(t)} + \kappa_3 \sum_{k=1}^{N} \frac{D_k(t)}{D(t)} S_k(t) \sigma^2 + \Omega_{t}\right].
  \end{align}
  Further, the convergence upper bound after $T$ rounds is
  \begin{align}
      & \mathbb E[F(w(T)|\mathcal D(T)) - F(w^*)] \notag \\ 
    = & \kappa_1^{T} \mathbb E[F(w(0)|\mathcal D(0))-F(w^*)] \notag \\
      & + \sum_{t=0}^{T-1}\kappa_1^{T-1-t} \left[\kappa_2 \frac{N\psi^2}{D(t)} + \kappa_3 \sum_{k=1}^{N} \frac{D_k(t)}{D(t)} S_k(t) \sigma^2 + \Omega_{t}\right].
  \end{align}
\end{proof}

\subsection{Proof of Proposition \ref{proposition:clientoptimalstrategy}}
\begin{proof}
    \label{proof:clientoptimalstrategy}
    For the long term optimization problem with dynamic constraint in Equ. (\ref{formulation:reformulated}), we construct a Hamilton equation $H_k(t)$ as follows:
    \begin{align}
      H_k(t) = & \frac{D_k(t)}{\phi(t)}R - \alpha_k {\Delta_k(t)}^2 - \beta_k {D_k(t)^2} \notag \\
      + & \lambda_k(t+1)\left((\theta-1) D_k(t) + \Delta_k(t)\right).
    \end{align}
    In addtion, $\frac{\partial^2 H_k(t)}{\partial {\Delta_k(t)}^2} = -2\alpha_k < 0$. That is, $H_k(t)$ is a concave function in $\Delta_k(t)$.
    To derive the optimal increment $\Delta_k(t)$ that minimize Equ. (\ref{formulation:reformulated}), it satisfies 
    \begin{align}
      \frac{\partial H_k(t)}{\partial \Delta_k(t)} & = 0, \\
      \frac{\partial H_k(t)}{\partial D_k(t)} & = \lambda_k(t) - \lambda_k(t+1).
    \end{align}
    By solving the above two formulas, we have
    \begin{align}
      \label{formulation:proof_delta}
      \Delta_k(t) & = \frac{1}{2\alpha_k} \lambda_k(t+1),\\
      \label{formulation:proof_lambda}
      \lambda_k(t) & = \theta \lambda_k(t+1) + \frac{R}{\phi(t)} - 2\beta_k D_k(t).
    \end{align}  

    In addition, it can be derived that $\Delta_k(T-1) = 0$ because $\Delta_k(T-1)$ decides round $T$'s increment for client $k$.
    The buffered data volume $D_k(T)$ will not be involved in the training rounds which ranges from 0 to $T-1$.
    Therefore $D_k(T)$ cannot bring any benefit for client $k$ under high collection expenditure.
    It's feasible for client $k$ to set $\Delta_k(T-1) = 0$ and stop data collection.
    Based on this, the boundary condition can be expressed as
    \begin{align}
      \label{formulation:boundary}
      \lambda_k(T-1) & = \frac{\partial \left(\frac{D(T-1)}{\phi(T-1)} R - \beta_k{D_k(T-1)}^2 \right)}{\partial D_k(T-1)} \notag \\
                     & = \frac{R}{\phi(T-1)}- 2\beta_k D_k(T-1).
    \end{align}
    Combing Equ. (\ref{formulation:proof_lambda}) with Equ. (\ref{formulation:boundary}), we have
    \begin{align}
      \lambda_k(t) = & \theta \lambda_k(t+1) + \frac{R}{\phi(t)} - 2\beta_k D_k(t) \notag \\
                  = & \theta^2 \lambda_k(t+2) \notag \\
                    & + \theta\left(\frac{R}{\phi(t+1)} - 2\beta_kD_k(t+1)\right) \notag \\
                    & + \frac{R}{\phi(t)} - 2\beta_kD_k(t) \notag \\
                  = & \cdots \notag \\
                  = & \theta^{T-1-t} \lambda_k(T-1) \notag \\
                    & + \sum_{\tau=0}^{T-2-t} \theta^{\tau} \left(\frac{R}{\phi(t + \tau)} - 2\beta_kD_k(t + \tau)\right) \notag \\
                  = & \theta^{T-1-t} \left(\frac{R}{\phi(T-1)} - 2\beta_k D_k(T-1)\right) \notag \\
                    & + \sum_{\tau=0}^{T-2-t} \theta^{\tau} \left(\frac{R}{\phi(t + \tau)} - 2\beta_k D_k(t + \tau)\right) \notag \\
                  = & \sum_{\tau = 0}^{T - 1 - t} \theta^{\tau} \left(\frac{R}{\phi(t + \tau)} - 2\beta_k D(t + \tau)\right) \notag \\
                  = & \sum_{\tau = t}^{T - 1} \theta^{\tau - t} \left(\frac{R}{\phi(\tau)} - 2\beta_k D_k(\tau)\right).
    \end{align}
    and 
    \begin{align}
      \label{formulation:ready_lambda}
      \lambda_k(t+1) = \sum_{\tau = t + 1}^{T - 1} \theta^{\tau - t - 1} \left(\frac{R}{\phi(\tau)} - 2\beta_k D_k(\tau)\right).
    \end{align}
    Substituting Equ. (\ref{formulation:ready_lambda}) into Equ. (\ref{formulation:proof_delta}), we obtain
    \begin{align}
      \label{formulation:ready_delta}
      \Delta_k(t) = \frac{1}{2 \alpha_k} \sum_{\tau = t + 1}^{T - 1} \theta^{\tau - t - 1} \left(\frac{R}{\phi(\tau)} - 2\beta_k D_k(\tau)\right).
    \end{align}
    Substituting Equ. (\ref{formulation:ready_delta}) into Equ. (\ref{formulation:update}), we obtain
    \begin{align}
      D_k(t+1) & = \theta D_k(t) + \Delta_k(t) \notag \\
            & = \theta^2 D_k(t-1) + \theta \Delta_k(t-1) + \Delta_k(t) \notag \\
            & = \cdots \notag \\
            & = \theta^{t+1} D_k(0) + \sum_{\tau = 0}^{t} \theta^{t-\tau} \Delta_k(\tau).
    \end{align}
    with $t \in \{0, T-2\}$ and $\Delta_k(T-1) = 0$.
  \end{proof}

  

\subsection{Proof of Proposition \ref{proposition:fixed point}}
\begin{proof}
  \label{proof:fixed_point}
  First, we construct a vector function. According to Equ. (\ref{formulation:datasize}), $D_k(t)$ is a function of $\Delta_k(\tau), \tau \in \{0,\cdots,t-1\}$. By inserting Equ. (\ref{formulation:delta}) into Equ. (\ref{formulation:datasize}), $D_k(t)$ is actually a function of $\phi(t)$ and $D_k(t), t \in \{0, \cdots, T-1\}$.
  We define this function as
  \begin{align}
    D_k(t) = \Psi_{k,t}( & \phi(0), \phi(1), \cdots, \phi(T-1), \notag\\
                         & D_k(0), D_k(1), \cdots, D_k(T-1)).
  \end{align}
  Furthermore, $\phi(t)$ is a function of $D_k(t), k \in \{1, \cdots, N\}$ at round $t$, so it can be derived that $D_k(t)$ is a function of $D_k(t), k \in \{1, \cdots, N\}, t \in \{0, \cdots, T-1\}$. That is
  \begin{align}
    \label{formulation:matrix}
    D_k(t) = \Psi_{k,t} & (D_1(0), D_1(1), \cdots, D_1(T-1), \notag \\
                        & D_2(0), D_2(1), \cdots, D_2(T-1), \notag \\
                        & \cdots, \notag \\
                        & D_N(0), D_N(1), \cdots, D_N(T-1)).
  \end{align}
  For ease of reading, we denote the parameter matrix in Equ. (\ref{formulation:matrix}) as $\boldsymbol{D}$. Then we have $D_k(t) = \Psi_{k,t}(\boldsymbol{D})$.
  To summarize buffered data volume $D_k(t)$ over the time horizon $t \in 
  \{0, \cdots, T-1\}$ and $k \in \{1, \cdots, N\}$, we have the following vector function as
  \begin{align}
         \Psi(\boldsymbol{D}) = (& \Psi_{1, 0}(\boldsymbol{D}), \Psi_{1, 1}(\boldsymbol{D}), \cdots, \Psi_{1, T-1}(\boldsymbol{D}) \notag \\
        & \Psi_{2, 0}(\boldsymbol{D}), \Psi_{2, 1}(\boldsymbol{D}), \cdots, \Psi_{2, T-1}(\boldsymbol{D}) \notag \\
        & \cdots, \notag \\
        & \Psi_{N, 0}(\boldsymbol{D}), \Psi_{N, 1}(\boldsymbol{D}), \cdots, \Psi_{N, T-1}(\boldsymbol{D})),
  \end{align}  
  Next, we examine whether fix point exists for $\Psi$. We bound $D_k(t)$ as [0, U], where $D_k(t) \geq 0$ means buffered data volume must be non-negative, and $D_k(t) \leq U$ means the storage capacity of buffer can't exceed $U$.
  The domain of $\Psi$ can be bounded as
  \begin{equation}
    \Pi = [0, U] \times [0, U] \times \cdots \times [0, U].
  \end{equation}
  Since $\Psi$ is a continuous mapping from $\Pi$ to $\Pi$, according to Brouwer's fixed point theorem, $\Psi$ has a fixed point in $\Pi$. Thus, there exists a fix point for $\boldsymbol{D}$. 
  Furthermore, there also exists a fix point for $\phi(t) = \sum_{k=1}^{N} D_k(t)$.
\end{proof}

\section{Additional Experiments}
\label{appendix:experiments}
\begin{figure}
  \centering

  \begin{subfigure}[b]{0.48\linewidth}
    \centering
    \includegraphics[width=\linewidth]{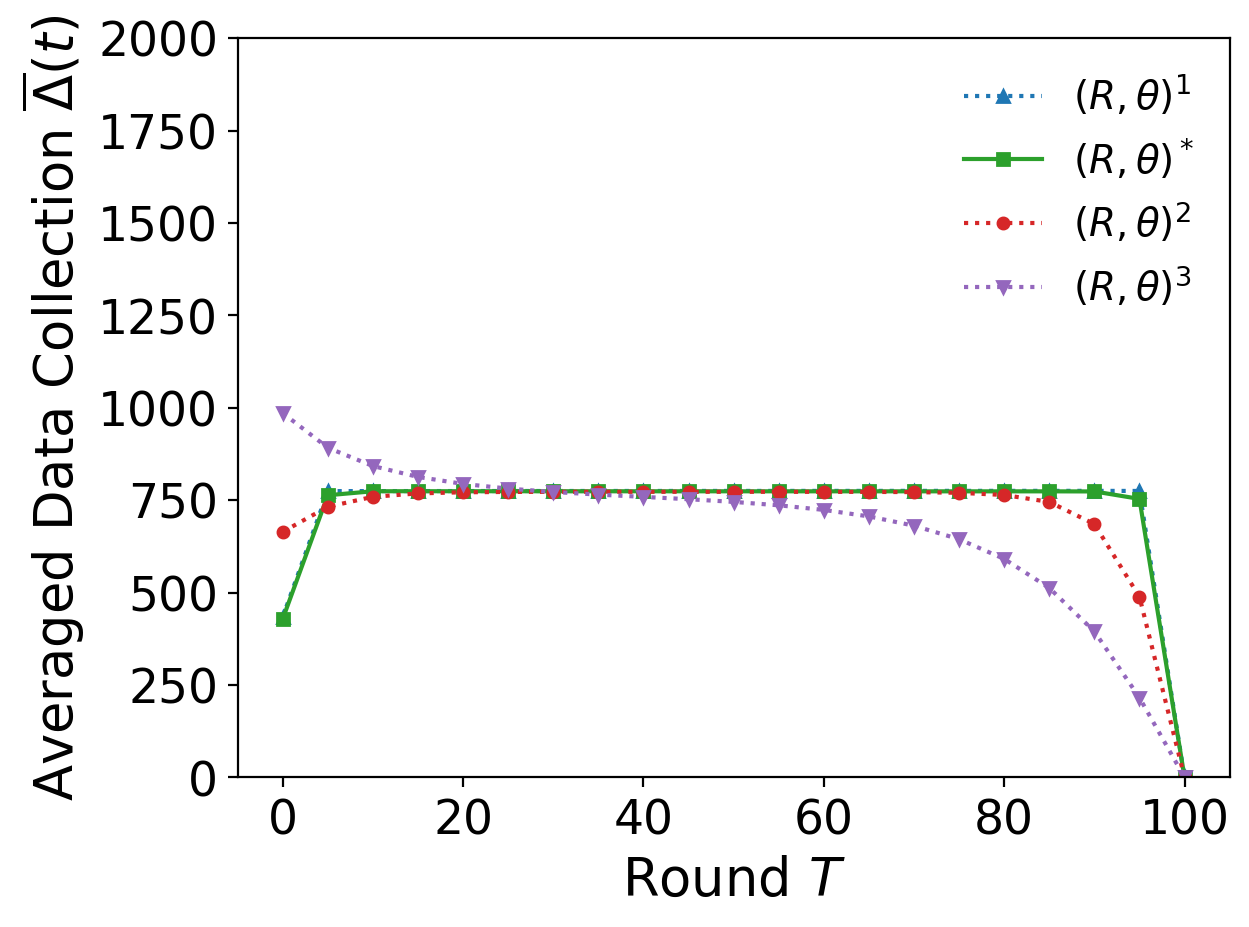}
    \caption{$\overline\Delta(t)$: $\theta$ changes}
  \end{subfigure}
  \hfill
  \begin{subfigure}[b]{0.48\linewidth}
    \centering
    \includegraphics[width=\linewidth]{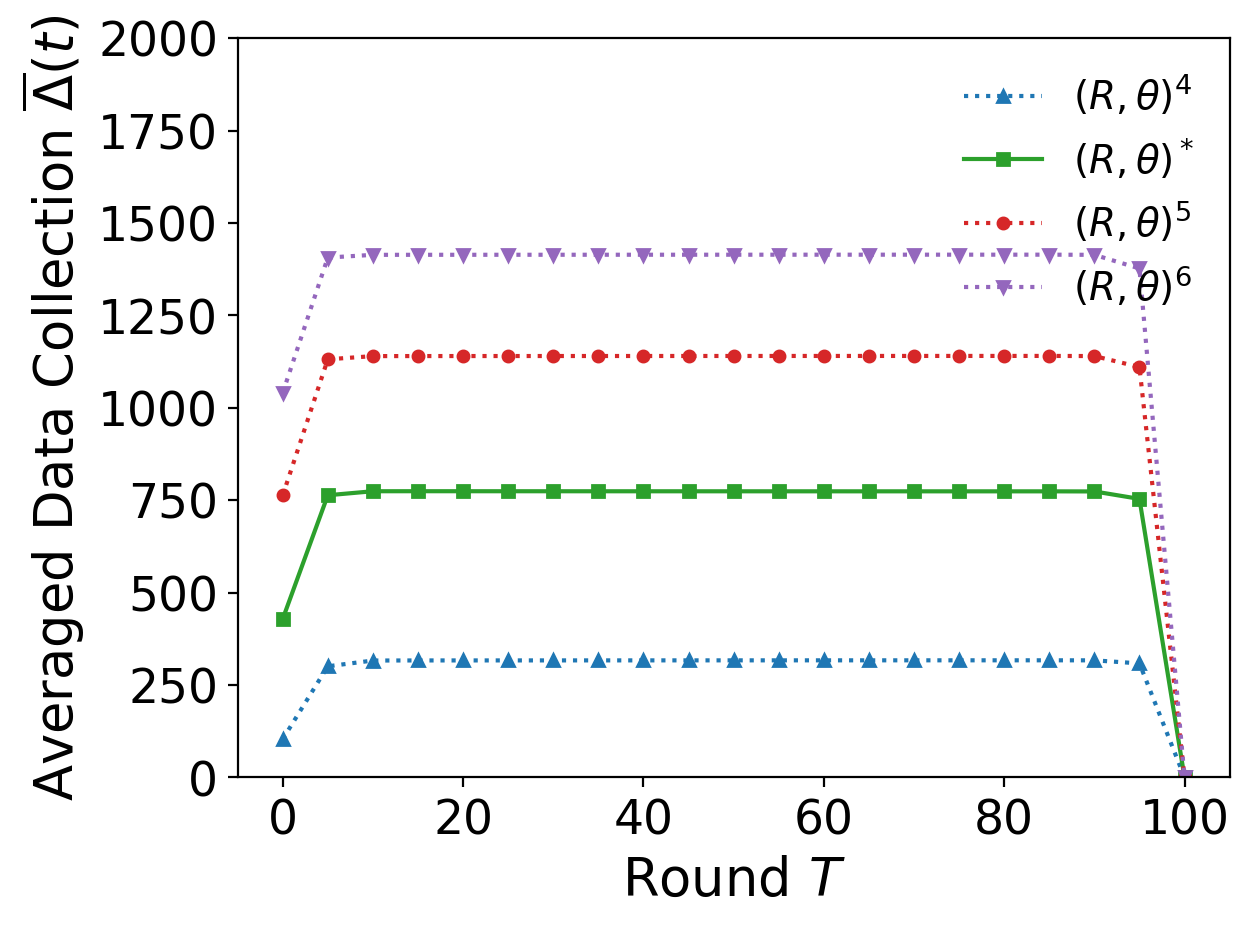}
    \caption{$\overline\Delta(t)$: $R$ changes}
  \end{subfigure}

  \vskip\baselineskip

  \begin{subfigure}[b]{0.48\linewidth}
    \centering
    \includegraphics[width=\linewidth]{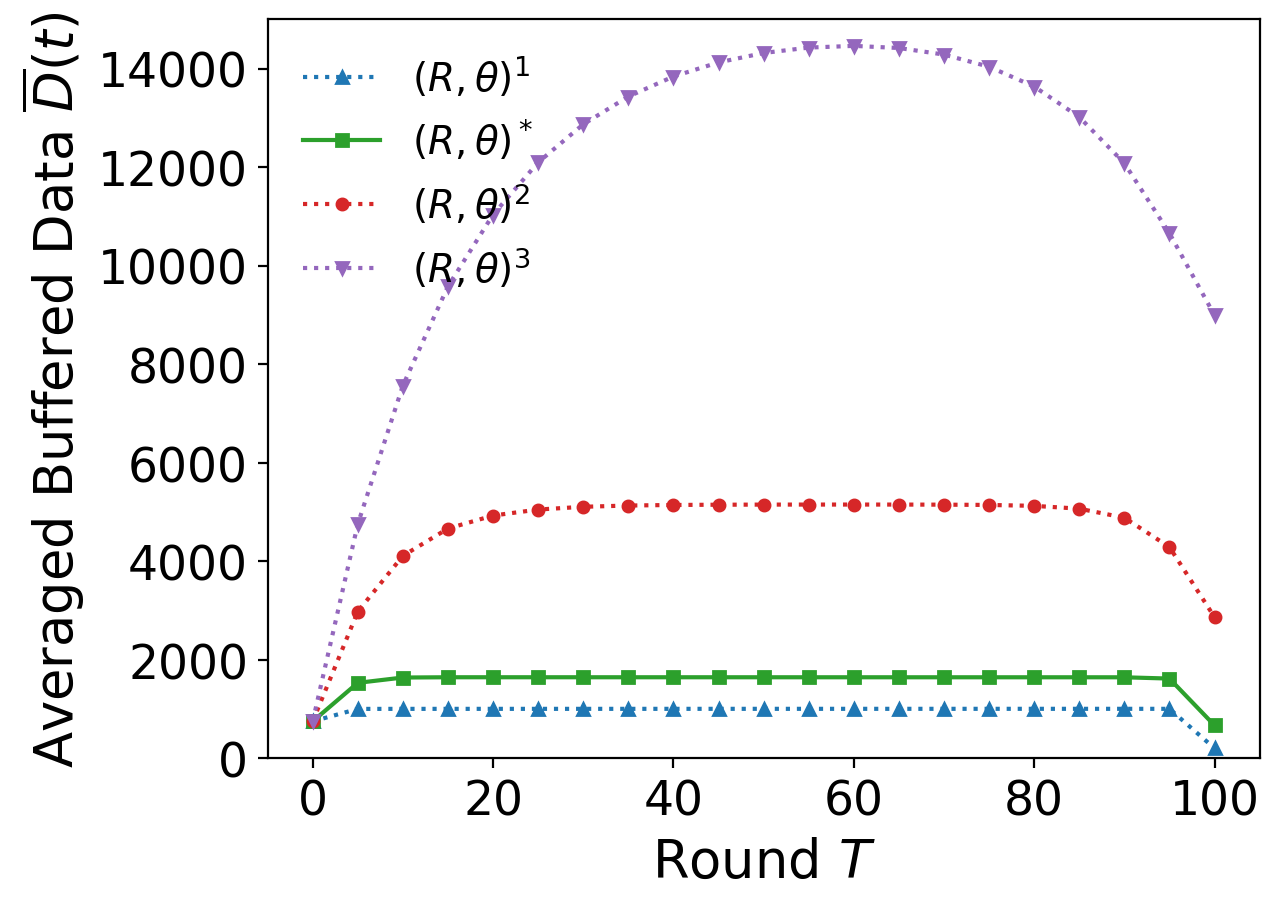}
    \caption{$\overline D(t)$: $\theta$ changes}
  \end{subfigure}
  \hfill
  \begin{subfigure}[b]{0.48\linewidth}
    \centering
    \includegraphics[width=\linewidth]{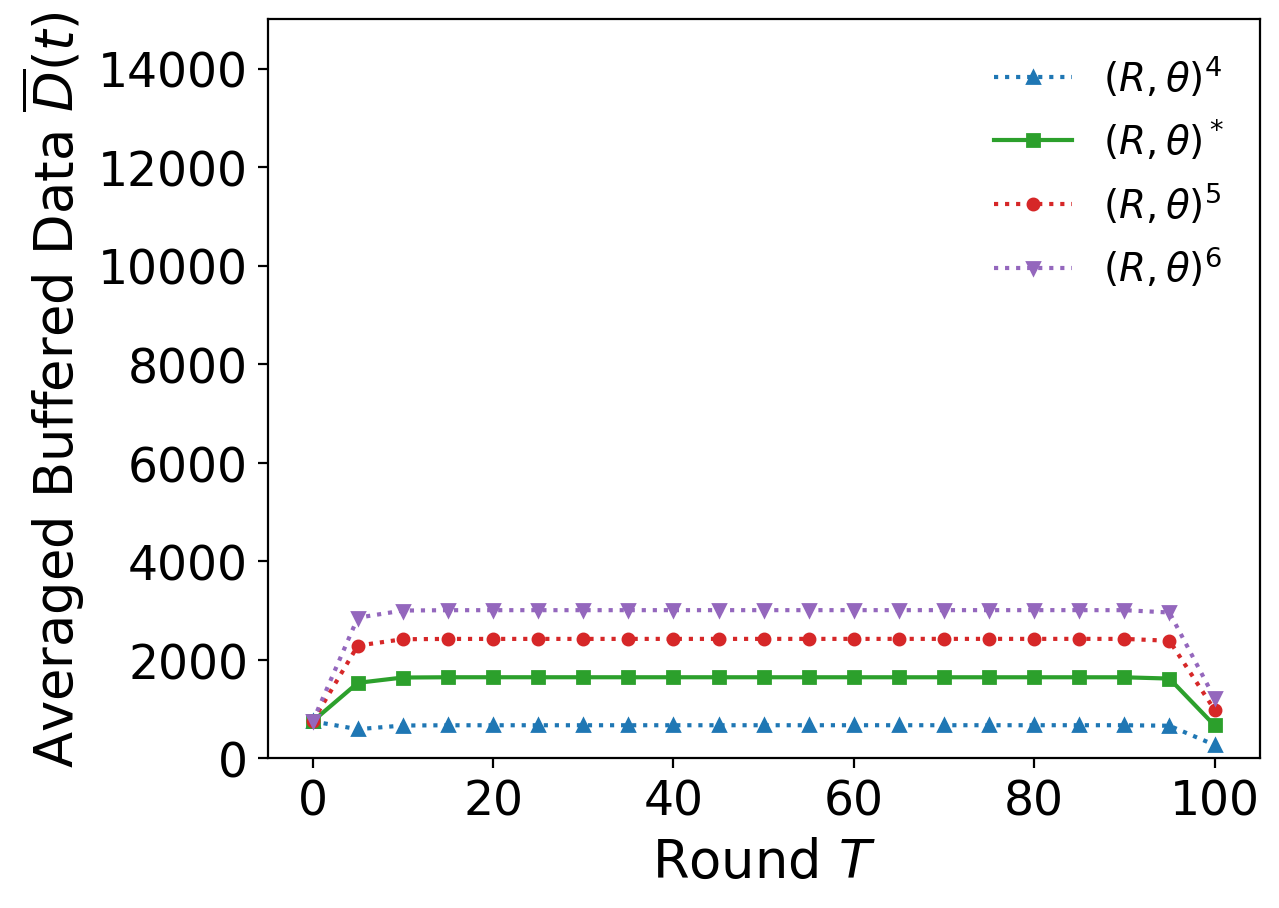}
    \caption{$\overline D(t)$: $R$ changes}
  \end{subfigure}

  \vskip\baselineskip

  \begin{subfigure}[b]{0.48\linewidth}
    \centering
    \includegraphics[width=\linewidth]{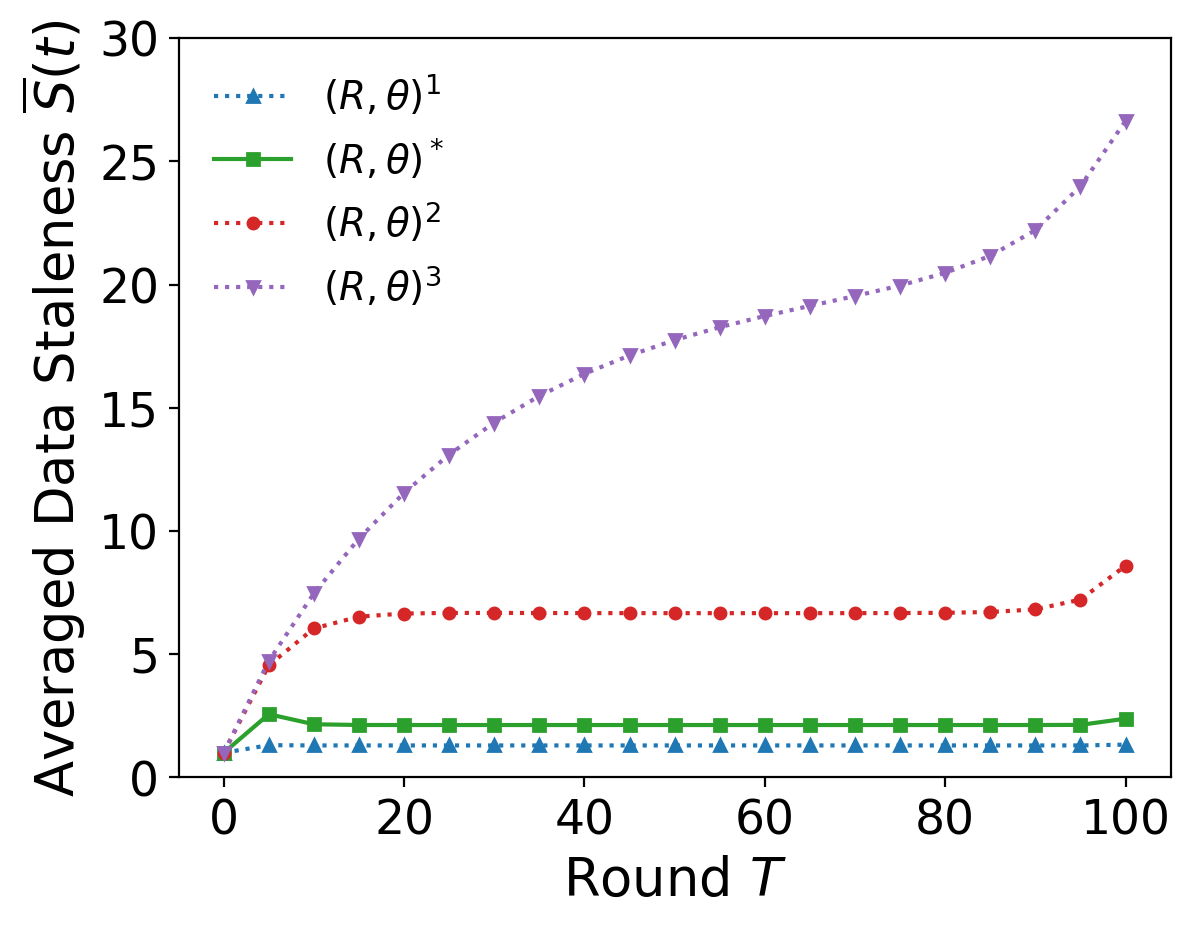}
    \caption{$\overline S(t)$: $\theta$ changes}
  \end{subfigure}
  \hfill
  \begin{subfigure}[b]{0.48\linewidth}
    \centering
    \includegraphics[width=\linewidth]{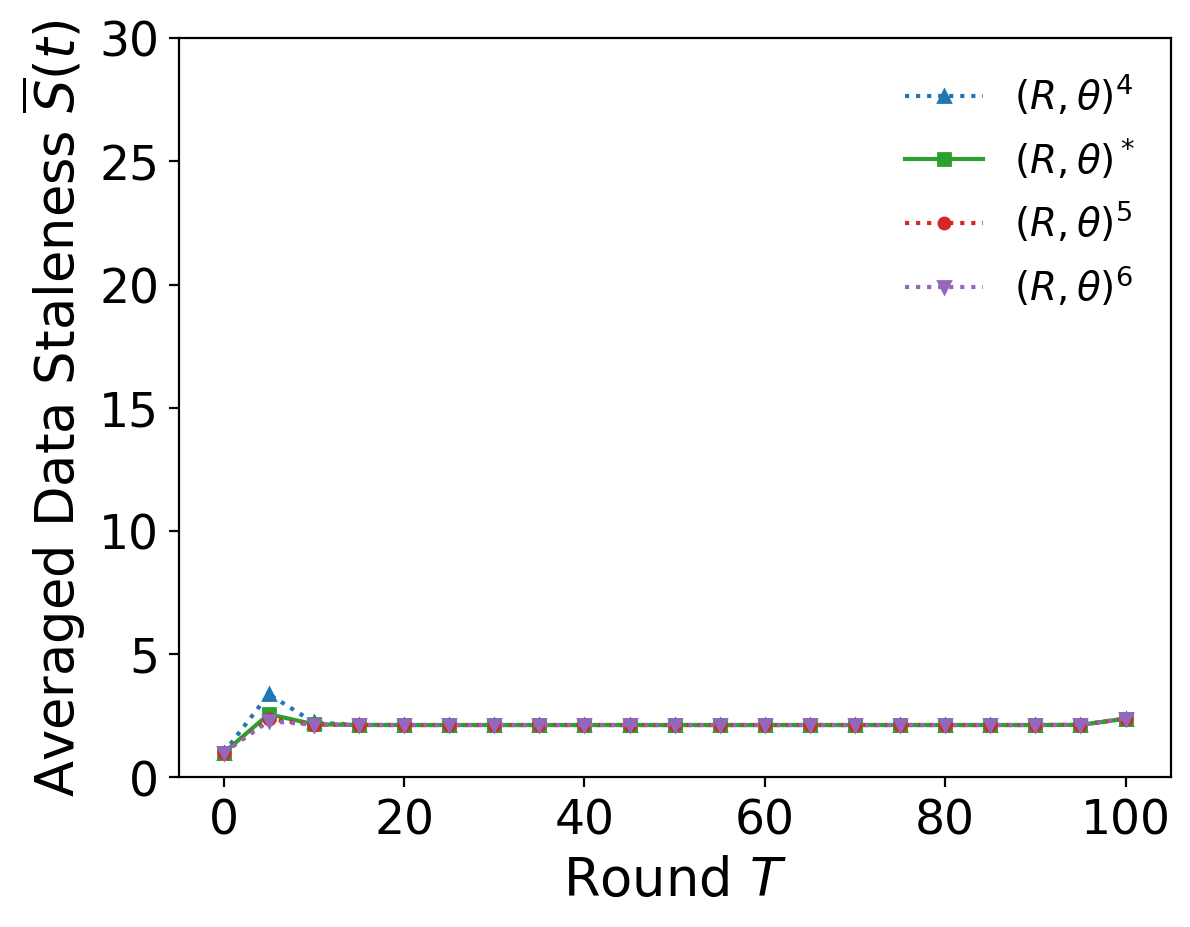}
    \caption{$\overline S(t)$: $R$ changes}
  \end{subfigure}
  
  \vskip\baselineskip

  \caption{Comparative analysis of averaged data collection volume $\overline\Delta(t)$, buffered volume$\overline D(t)$, and staleness $\overline S(t)$ over communication rounds $T$ under various server strategies.}
  \vspace{1.5em}
  \label{fig:appendix}
\end{figure}

\textbf{Effect of Server's Strategy on Client's Data Update Strategy: }
In this section, we investigate how the server’s strategy \((R, \theta)\) influences client-side data updates across three metrics: averaged data collection volume \(\overline\Delta(t) = \frac{1}{N}\sum_{k=1}^N \Delta_k(t)\), buffered data volume \(\overline D(t) = \frac{1}{N}\sum_{k=1}^N D_k(t)\), and data staleness \(\overline S(t) = \frac{1}{N}\sum_{k=1}^N S_k(t)\), as illustrated in Figure~\ref{fig:appendix}. We reuse the experimental settings from the main text, including the baseline strategy \((R, \theta)^*\), \((R, \theta)^{1\text{-}3}\)($R$ changes) and \((R, \theta)^{4\text{-}6}\)($\theta$ changes).

From Figure~\ref{fig:appendix}(a)(b), we observe that with fixed \(R\), a higher \(\theta\) leads to a larger \(\overline\Delta(t)\) at early stages, since a greater retention rate allows previously collected data to persist in the buffer and contribute to the reward longer. However, in later rounds, clients reduce data collection to save costs, resulting in smaller \(\overline\Delta(t)\) under large \(\theta\). In contrast, increasing \(R\) consistently boosts \(\overline\Delta(t)\), as greater incentives encourage clients to collect more fresh data. Notably, \(\overline\Delta(t)\) converges to zero before training ends, consistent with the boundary behavior in Eq.~(\ref{formulation:delta}).
Figure~\ref{fig:appendix}(c)(d) shows that \(\overline D(t)\) grows with both \(R\) and \(\theta\), as larger \(R\) brings more new data, and larger \(\theta\) retains more old data. Moreover, the stabilization of \(\overline D(t)\) aligns with the trend of \(\overline\Delta(t)\), reflecting the inflow–retention balance.
In addition, Figure~\ref{fig:appendix}(e)(f) reveals that staleness \(\overline S(t)\) is mainly influenced by \(\theta\). As \(\theta\) increases, outdated data accumulates, driving \(\overline S(t)\) higher. Meanwhile, \(R\) has little effect on staleness, as it does not directly control retention. These results suggest that \(\theta\) plays the key role in balancing data volume and freshness for better performance.

\end{document}